%% file: main_arxiv.tex
\documentclass[10pt,twocolumn,letterpaper]{article}

\usepackage{wacv}              

\definecolor{wacvblue}{rgb}{0.21,0.49,0.74}
\usepackage[pagebackref,breaklinks,colorlinks,allcolors=wacvblue]{hyperref}

\usepackage[accsupp]{axessibility} 

\usepackage{algorithm}
\usepackage{amsmath}
\usepackage{amsfonts}
\usepackage{amssymb}
\usepackage{bm}
\usepackage{multirow}
\usepackage{array}
\usepackage{ragged2e}
\usepackage{booktabs}
\usepackage{subcaption}
\usepackage{algpseudocode}
\newcolumntype{R}[1]{>{\RaggedLeft\arraybackslash}m{#1}}
\newcolumntype{L}[1]{>{\RaggedRight\arraybackslash}m{#1}}
\newcolumntype{C}[1]{>{\center\arraybackslash}m{#1}}

\title{\textsc{AD}$^2$: Analysis and Detection of Adversarial Threats in Visual Perception for End-to-End Autonomous Driving Systems}

\author{
        Ishan Sahu\textsuperscript{1, 2},\hspace{2em} Somnath Hazra\textsuperscript{1},\hspace{2em} Somak Aditya\textsuperscript{1},\hspace{2em} Soumyajit Dey\textsuperscript{1}\\
    \textsuperscript{1} Indian Institute of Technology Kharagpur\hspace{2em}
    \textsuperscript{2} TCS Research, India\\
    {\tt\small \{ishan.sahu@kgpian, somnathhazra@kgpian, saditya@cse, soumya@cse\}.iitkgp.ac.in }
}

\begin{document}

\maketitle

\begin{abstract}
End-to-end autonomous driving systems have achieved significant progress, yet their adversarial robustness remains largely underexplored. In this work, we conduct a closed-loop evaluation of state-of-the-art autonomous driving agents under black-box adversarial threat models in CARLA. Specifically, we consider three representative attack vectors on the visual perception pipeline: (i) a physics-based blur attack induced by acoustic waves, (ii) an electromagnetic interference attack that distorts captured images, and (iii) a digital attack that adds ghost objects as carefully crafted bounded perturbations on images. Our experiments on two advanced agents, Transfuser and Interfuser, reveal severe vulnerabilities to such attacks, with driving scores dropping by up to 99\% in the worst case, raising valid safety concerns. To help mitigate such threats, we further propose a lightweight Attack Detection model for Autonomous Driving systems (\textsc{AD}$^2$) based on attention mechanisms that capture spatial–temporal consistency. Comprehensive experiments across multi-camera inputs on CARLA show that our detector achieves superior detection capability and computational efficiency compared to existing approaches.
\end{abstract}

\section{Introduction}
\label{sec1}

Autonomous driving (AD) systems are rapidly progressing from research prototypes to real-world deployment. Despite continuous progress, road safety still faces severe risks, and well-designed AD systems could reduce over 1.2 million fatalities reported annually \cite{worldeconomicforumAutonomousVehiclesTimeline2025}. These systems promise to improve transportation efficiency and reduce accidents, yet their safety and reliability remain a fundamental concern \cite{chouguleComprehensiveReviewLimitations2024,national2017automated}. Current deployments are typically restricted to specific locations or operating conditions, reflecting both safety issues and the unresolved technical limitations of AD technology. A critical requirement for improving AD adoption is robustness: systems must perform reliably not only under nominal conditions but also under out-of-distribution scenarios introduced by environmental perturbations or adversarial attacks \cite{fingscheidtDeepNeuralNetworks2022}.

Recent studies in machine learning have revealed the fragility of neural networks to adversarial perturbations \cite{szegedyIntriguingPropertiesNeural2014,vassilevAdversarialMachineLearning2024}. Addition of hardly perceptible perturbations to inputs can lead to significant mispredictions. Similar vulnerabilities have been observed in perception modules of autonomous vehicles, such as localization of the vehicle or detection of static objects \cite{kim2024survey,badjieAdversarialAttacksCountermeasures2024,chiAdversarialAttacksAutonomous2024,guesmiNavigatingThreatsSurvey2024,nguyenSurveyEvaluationAdversarial2025}. However, most of the prior work emphasises on adversarial robustness of one of the subtasks and modules in isolation rather than considering the overall end-to-end driving performance. As a result, existing analyses fail to capture how perturbations propagate through the entire control pipeline and ultimately affect driving safety.

\begin{figure}[!t]
    \centering
    \includegraphics[width=0.98\linewidth]{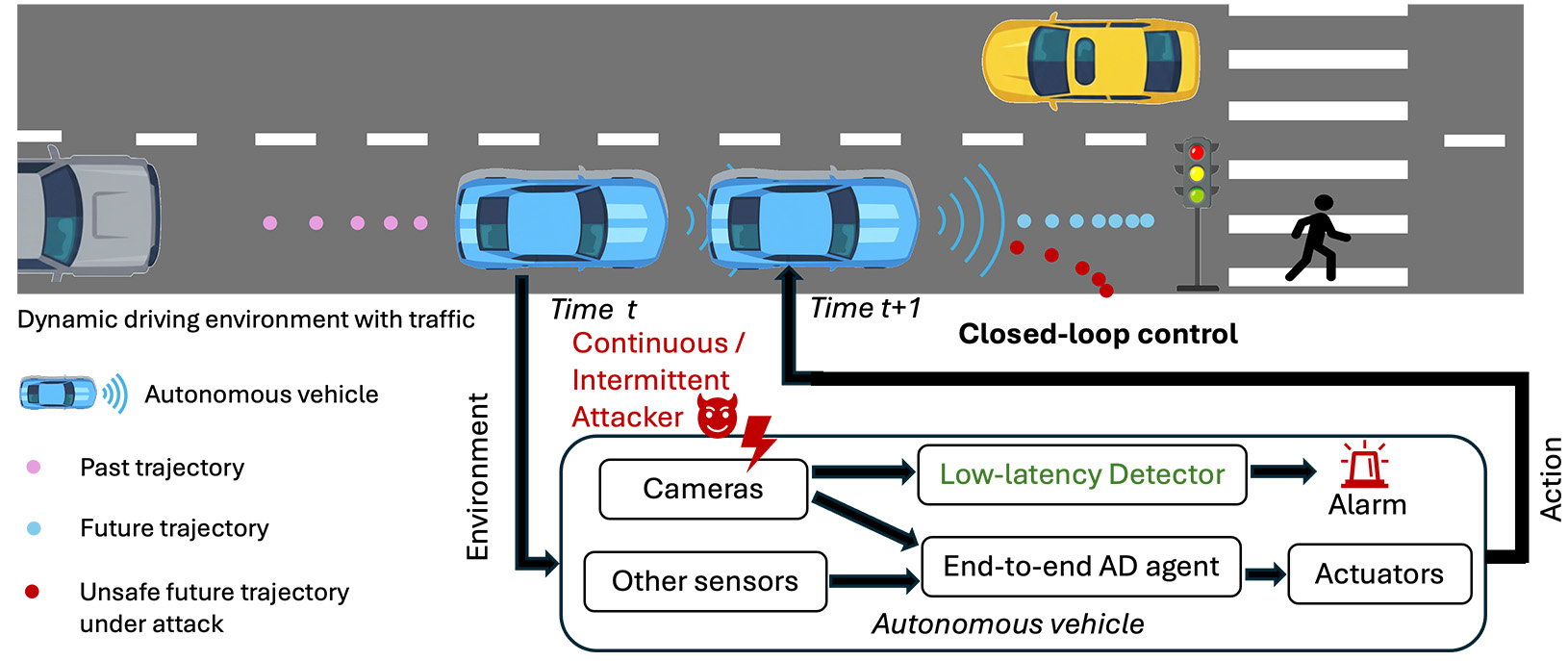}
    \caption{Illustration of an adversarial attack on the perception system and a low-latency detector to protect against the attack in an end-to-end closed system model for autonomous driving.}
    \label{fig1}
\end{figure}

The design of AD systems have evolved from a modular approach towards end-to-end approaches. Unlike modular  systems, which rely on different independently trained modules or subsystems, \emph{end-to-end agents} directly map raw sensory inputs to control signals or trajectories \cite{chenEndtoendAutonomousDriving2024}. This design enables joint optimization of perception and planning to create motion plans and then control actions through a controller. However, the coupling of perception and decision-making also introduces new challenges: internally the submodels communicate using latent features and adversarial perturbations on sensor inputs may cause nonlinear effects on the overall driving behavior. Their lack of transparency (black box models) makes it challenging to understand how a perturbation affects the decision-making process. Moreover, the \emph{closed-loop} nature of AD, where outputs at one step affect future inputs \cite{li2025hydra}, can lead to cascading errors under attack. Despite its importance, adversarial robustness of end-to-end AD in closed-loop settings has received very limited investigation.

In this work, we address this gap by performing a \emph{study of adversarial robustness in state-of-the-art end-to-end driving systems}. We focus on threat models targeting the visual perception pipeline from cameras, since cameras are among the primary perception sensors in AD systems. We design three representative attack vectors: (i) a physics-based blur attack caused by acoustic interference, (ii) an electromagnetic interference attack distorting captured images, and (iii) a digital attack injecting ghost objects through bounded perturbations. Importantly, while cameras are under attack, other sensors remain unaffected, enabling us to assess the sensitivity of agents to visual perception alone.

Motivated by these findings, we further propose an \emph{Adversarial \underline{A}ttack \underline{D}etection framework for end-to-end \underline{A}utonomous \underline{D}riving (\textsc{AD}$^2$)}. Unlike the existing methods that require access to internal intermediate outputs or modification of the underlying driving agents, our detector operates externally and with low computational overhead, making it suitable for practical deployment. Experiments on CARLA simulator \cite{dosovitskiyCARLAOpenUrban2017} confirm that our approach outperforms baselines both in detection accuracy and computational efficiency. In summary, our contributions are:
\begin{itemize}
    \item We conduct a closed-loop adversarial robustness analysis of state-of-the-art end-to-end autonomous driving agents under black-box attacks targeting perception.
    \item We show that intermittent perturbations are sufficient to induce catastrophic failures, demonstrating the fragility of AD.
    \item We propose an adversarial attack detection model, \textsc{AD}$^2$ for identifying erroneous frames from the camera sensor, which is effective without requiring modifications of the AD agents or access of internal latent outputs.
    \item Experiments on the CARLA simulator validate both the severity of the existing adversarial threats and the effectiveness of our detection framework.
\end{itemize}

\section{Related Work}
\label{sec2}

We briefly discuss existing related work organized into following subcategories.

\subsection{End-to-End Autonomous Driving Systems}

Autonomous driving systems are often categorized as modular or end-to-end. Conventional systems separate perception, prediction, and planning, each with complete intermediate outputs, but suffer from error propagation and lack of joint optimization. In contrast, end-to-end systems are fully differentiable pipelines that map raw sensory inputs directly to control signals, enabling joint training and global optimization \cite{chenEndtoendAutonomousDriving2024,ozaibiEndtoendAutonomousDriving2024}. Recent models such as Transfuser \cite{chittaTransFuserImitationTransformerbased2023}, Interfuser \cite{shaoSafetyenhancedAutonomousDriving2023}, and ReasonNet \cite{shaoReasonNetEndtoendDriving2023} have demonstrated strong performance in the CARLA leaderboard 1.0 challenge.

\subsection{Adversarial Attacks on AD Systems}

Most existing attack studies focus on modular systems, targeting subtasks such as object detection, classification, or lane detection \cite{zhangEvaluatingAdversarialAttacks2022,satoWIPRobustnessLane2022}. Physical-world attacks on perception sensors are also well documented \cite{chiAdversarialAttacksAutonomous2024}. In contrast, adversarial robustness of end-to-end systems has only recently been explored. For example, \cite{wuAdversarialDrivingAttacking2023} showed lane deviation under white-box perturbations on a simplified model, and \cite{wangAttackEndtoendAutonomous2024} proposed noise injection across submodules. However, these studies often assume unrealistic white-box access and lack evaluation on state-of-the-art multi-modal systems.

\subsection{Detection of Adversarial Attack}

Detection aims to distinguish adversarial from benign inputs using statistical tests, reconstruction errors, or auxiliary classifiers \cite{metzenDetectingAdversarialPerturbations2017,feinmanDetectingAdversarialSamples2017,xuFeatureSqueezingDetecting2018,fangKernelPCAOutofdistribution2024}. While stronger adaptive attacks are possible against such detectors \cite{carliniAdversarialExamplesAre2017,tramerDetectingAdversarialExamples2022}, such attacks may not be feasible in terms of computational complexity and threat model of practical real-time systems. Recent works have considered patch detection \cite{liuSegmentCompleteDefending2022} and spatio-temporal consistency \cite{manThatPersonMoves2023,xuPhyScoutDetectingSensor2024}, but often rely on intermediate outputs such as bounding boxes or classification scores, which are unavailable in end-to-end pipelines. Other approaches such as CyberDet \cite{cyberdet} exploit handcrafted differences on inputs but remain limited to image classification settings.

In summary, most prior studies either evaluate adversarial attacks on isolated perception modules or design detection methods requiring intermediate outputs. Research on adversarial robustness and detection for end-to-end AD systems with multi-modal perception is still limited. Our work addresses this gap by (i) discussing the efficacy of existing attacks using closed-loop evaluation on state-of-the-art end-to-end systems, and (ii) developing a detection method that leverages spatial and temporal consistency without relying on intermediate task-specific outputs.

\section{Adversarial Robustness against Black-Box Attacks for End-to-End AD Systems}
\label{sec3}

\subsection{Adversarial Attacks on AD Systems}

We focus on perturbations to camera-based perception in end-to-end AD systems. Formally, let $\mathbf{x}_0, \mathbf{x}_1, \dots, \mathbf{x}_T \in \mathbb{R}^m$ denote the sequence of sensor inputs  at time instances $t = 0, \dots, T$; and $\mathbf{y}_0, \mathbf{y}_1, \dots, \mathbf{y}_T$ denote the corresponding control outputs of an autonomous driving system $A$. An adversarial attack is a mapping $\Psi:\mathbb{R}^m \to \mathbb{R}^m$ such that the perturbed inputs $\mathbf{x}^{\text{adv}}_t = \Psi(\mathbf{x}_t)$, $t=0,\dots,T$, induce a sequence of control signals $\mathbf{y}^{\text{adv}}_t$ that disrupt safe driving behaviour.

Our attack model assumes a restricted black-box setting. The adversary cannot query the target model or access its parameters, but may optimize perturbations using surrogate models or apply predefined manipulations. Attacks are untargeted, aiming to disrupt the normal driving behaviour rather than enforcing specific outcomes. Adversarial perturbations can be broadly categorized as:
(a) Digital attacks, where perturbations are directly added to the digital input images (e.g., FGSM \cite{goodfellowExplainingHarnessingAdversarial2015}, PGD \cite{madryDeepLearningModels2018}),
(b) Physical attacks, where objects in the environment are modified or created with specific features (e.g., DARTS \cite{sitawarinDartsDeceivingAutonomous2018}),
(c) Physics-based attacks, exploiting hardware vulnerabilities (e.g., acoustic or electromagnetic interference \cite{jiPoltergeistAcousticAdversarial2021, tpatch,GlitchHiker,liao2025your}).
In this work, we evaluate three representative attacks covering both digital and physics-based categories.
\begin{enumerate}
    \item Poltergeist \cite{jiPoltergeistAcousticAdversarial2021}: an acoustic attack that manipulates inertial sensors of image stabilizers, leading to blurred images. We simulate this effect by applying blur transformations to camera inputs.
    \item Steal Now, Attack Later (SNAL) \cite{chenStealNowAttack2024}: a digital attack projecting ghost objects from pre-collected data into input images. Perturbations are colour-manipulated to remain within an $\epsilon$-ball, overwhelming object detectors with false positives.
    \item Electromagnetic Signal Injection Attack (ESIA) \cite{liao2025your}: an interference-based attack injecting adversarial signals into camera circuits, producing image distortions with coloured strips between the image capture and processing units. We adopt the simulation method of \cite{liao2025your} to generate such patterns.
\end{enumerate}
This setup allows us to examine the impact of both hardware-induced distortions and adversarially crafted digital perturbations on end-to-end AD systems.

\subsubsection{Problem Formulation}

Referring to the previous description of adversarial attack, consider an AD system $A$ that receives multimodal sensor inputs $\mathbf{x}_t = [\mathbf{x}^{C_1:C_n}, \mathbf{x}^L, \mathbf{x}^R, \mathbf{x}^G, \mathbf{x}^U, \mathbf{x}^S ]_t$ at time $t$, where $C_i$ are $n$ cameras, $L$ is lidar, $R$ radar, $G$ Global Navigation Satellite System (GNSS), $U$ Inertial Measurement Unit (IMU), and $S$ speedometer. The system outputs control signals $\mathbf{y}_t = [y^{\text{steer}}, y^{\text{throttle}}, y^{\text{brake}}]_t$:
\begin{equation}
\mathbf{y}_t \leftarrow A(\mathbf{x}_t).
\end{equation}
Under attack, only the camera inputs are perturbed by adversarial attack $\Psi$: $\mathbf{x}_t^\text{adv} = [ \mathbf{x}^{C_1:C_n,\text{adv}}, \mathbf{x}^L, \mathbf{x}^R, \mathbf{x}^G, \mathbf{x}^U, \mathbf{x}^S ]_t$, which leads to adversarial control signals:
\begin{equation}
\mathbf{y}_t^\text{adv} \leftarrow A(\mathbf{x}_t^\text{adv}).
\end{equation}

\begin{figure}[h]
	\centering
	\includegraphics[width=\linewidth]{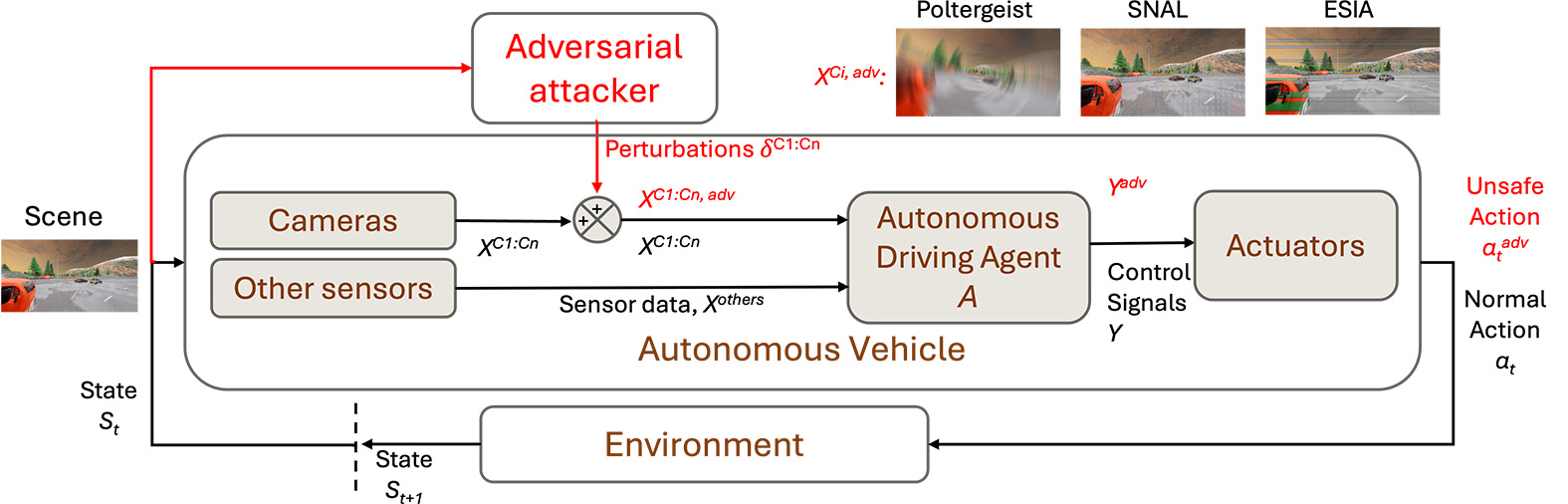}
	\caption{Adversarial attack setup for evaluation of end-to-end autonomous driving systems. 
    Red marked labels are used to indicate the change when the system is under attack.}
	\label{fig:attack_setup}
\end{figure}

The perturbations are generated by adversary under restricted black box threat model based on some predefined criteria or use some related surrogate model to optimize its attack. 
This setup is depicted in \Cref{fig:attack_setup}. Attacks may be applied continuously at every timestep, or intermittently at intervals of $d$ steps. This allows us to evaluate both sustained and sparse perturbations in closed-loop driving.

Before moving onto the results that highlight the performance of the studied adversarial attacks, we explain an attack detection mechanism that is specifically curated for such end-to-end AD systems.

\subsection{Adversarial Attack Detection}

The previous section formalized adversarial perturbations on end-to-end AD systems. While such perturbations can severely degrade driving performance, their impact can be mitigated if the system is able to detect adversarial inputs online. This motivates the following detection problem.

\subsubsection{Problem Formulation} 

At each timestep $t$, the AD system receives multi-camera inputs $\mathbf{x}^C_t$ along with the previous frame $\mathbf{x}^C_{t-1}$. The goal of a detector $D$ is to assign a label $l^{\mathbf{x}_t}$ among the \emph{four classes}: benign or one of the three attack categories described before, Poltergeist, SNAL or ESIA:
\begin{equation}
l^{\mathbf{x}_t} \leftarrow D(\mathbf{x}_{t-1}, \mathbf{x}_t).
\label{eqn3}
\end{equation}

\begin{figure}[h!]
	\centering
    \begin{subfigure}[b]{0.41\textwidth}
        \centering
        \includegraphics[width=\textwidth]{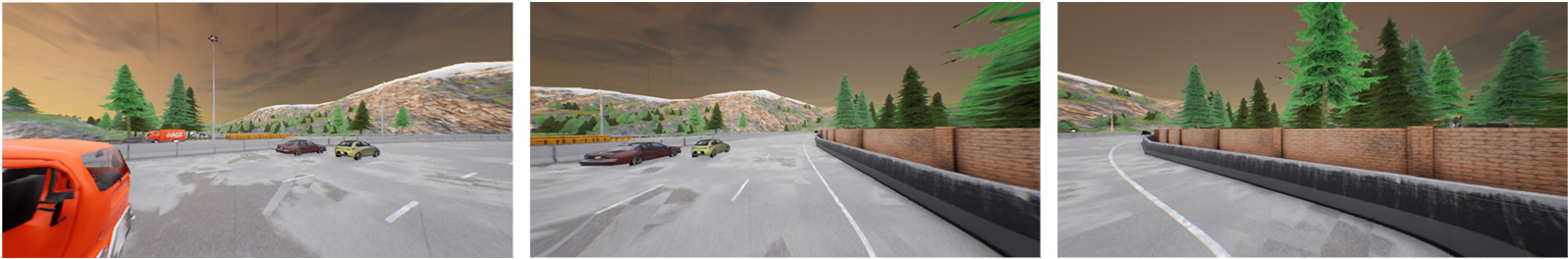}
        \caption{Input images from left, centre and right cameras respectively when there is no attack.}
        \label{fig:normal}
    \end{subfigure}
    \hfill
    \begin{subfigure}[b]{0.41\textwidth}
        \centering
        \includegraphics[width=\textwidth]{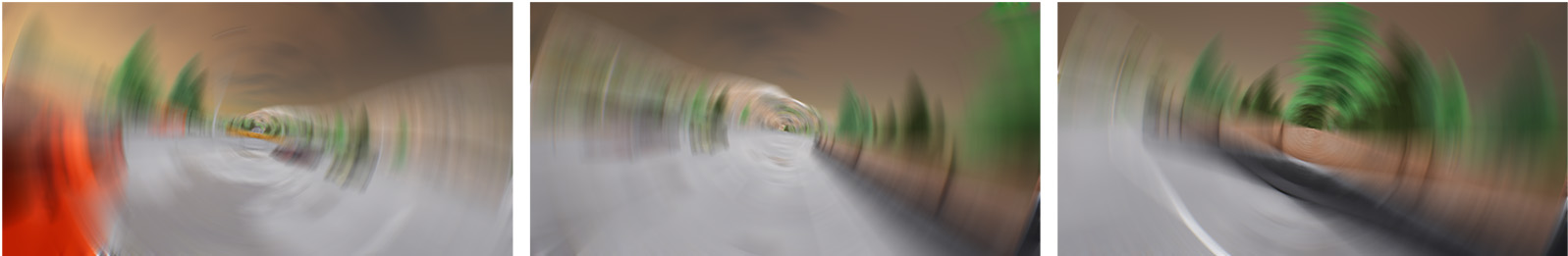}
        \caption{Images from left, centre and right cameras under Poltergeist.}
        \label{fig:polter}
    \end{subfigure}
    \hfill
    \begin{subfigure}[b]{0.41\textwidth}
        \centering
        \includegraphics[width=\textwidth]{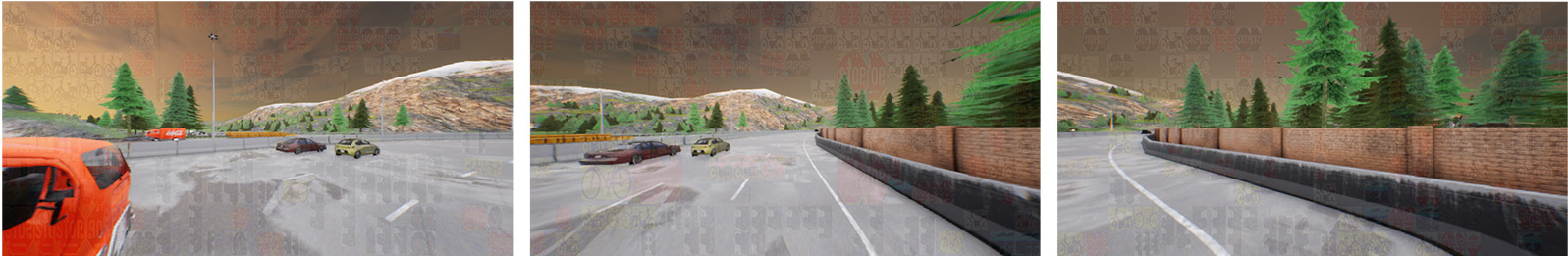}
        \caption{Images from left, centre and right cameras under SNAL.}
        \label{fig:snal}
    \end{subfigure}
    \begin{subfigure}[b]{0.41\textwidth}
        \centering
        \includegraphics[width=\textwidth]{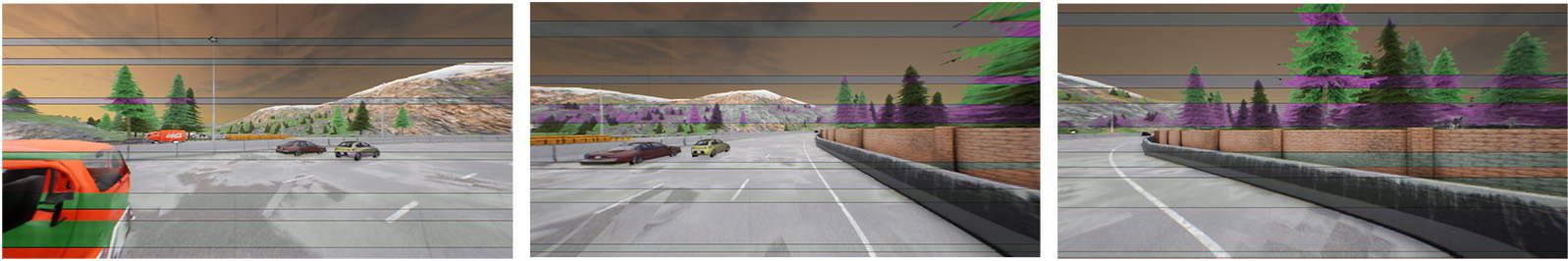}
        \caption{Images from left, centre and right cameras under ESIA.}
        \label{fig:esia}
    \end{subfigure}
    \caption{Example images from cameras of Transfuser. The 3 cameras in Transfuser, Interfuser have overlapping fields of view.}
    \label{fig:example_input_images}
\end{figure}

We further investigate whether adversarially perturbed images can be detected as out-of-distribution (OOD) anomalies, without explicitly learning the perturbation characteristics. \Cref{fig:example_input_images} illustrates typical benign inputs and inputs corrupted by Poltergeist, SNAL, and ESIA attacks. Here, the perturbed images in \Cref{fig:polter}, \Cref{fig:snal}, and \Cref{fig:esia} are being considered OOD with respect to benign images in \Cref{fig:normal}. We will revisit these inputs while describing the detection model that we have used.

\subsubsection{Motivation and Intuition behind the Method}

Adversarial perturbations often violate the \emph{consistency} in multi-sensor perception streams. We propose to learn a supervised classification model to capture consistency in image inputs:  
(i) \textbf{Spatial consistency}, referring to the identification of irregularities caused by adversarial perturbations using overlapping fields from multiple cameras. This is based on consistency in environment perception from multiple cameras.  
(ii) \textbf{Temporal consistency}, referring to coherence in perception across consecutive time-steps under normal driving conditions. We expect that, depending on the sampling rate of the sensor, within a small time interval there would be aspects of consistency.

By jointly modeling these consistencies, we aim to identify adversarial perturbations that disrupt either spatial alignment or temporal evolution. Our approach has the following novel design elements.
\begin{enumerate}
    \item Use of muti-camera inputs with overlapping fields of view to ascertain consistency in spatial perception. The information is processed using transformer based module on feature representations from three cameras capturing left, right and centre fields of view.
    \item Use of visual perception inputs from successive timesteps to check for temporal consistency. The information is concatenated from the three camera modules for two successive timesteps and processed through a temporal transformer module.
\end{enumerate}
We will now explain each element from our proposed architecture in the following sub-section.

\subsubsection{\textsc{AD}$^2$: A Real-time Adversarial Attack Detector for AD systems}

\begin{figure*}[h!]
	\centering
    \includegraphics[width=0.95\linewidth]{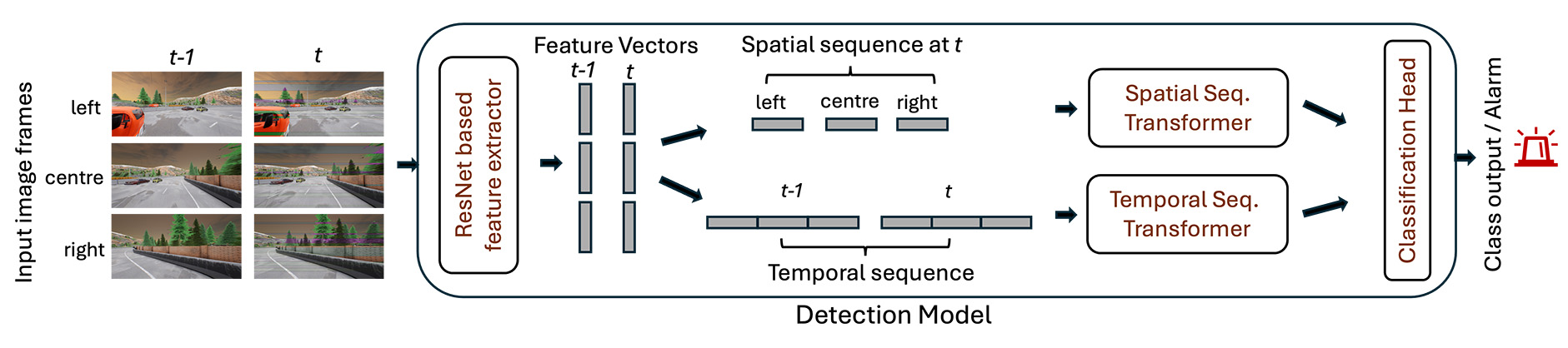}
	\caption{Proposed detection model, \textsc{AD}$^2$, leveraging spatial and temporal consistency.}
	\label{fig:model_architecture}
\end{figure*}

Our detector, $D$, is a deep neural network that integrates convolutional feature extraction with transformer-based reasoning modules (\Cref{fig:model_architecture}). The pipeline consists of:  
(1) a feature extraction backbone;  
(2) transformer blocks for spatial and temporal consistency modelling; and  
(3) a classification head for predicting $l^{\mathbf{x}_t} \in \{\text{benign}, \text{Poltergeist}, \text{SNAL}, \text{ESIA}\}$.

\textbf{Input and Feature Extraction.}
As input, we consider image vectors $[\mathbf{x}^C_{t-1}, \mathbf{x}^C_{t}]$, where $\mathbf{x}^C_{t} = [\mathbf{x}_t^{C_\text{left}}, \mathbf{x}_t^{C_\text{centre}}, \mathbf{x}_t^{C_\text{right}}]$ are RGB images captured from three cameras. Each image is processed by a ResNet \cite{He_2016_CVPR} backbone with two residual blocks and a linear layer, producing a compact feature vector $\mathbf{v}_t \in \mathbb{R}^{64}$ corresponding to each $\mathbf{x}_t$. ResNet is chosen as the feature extractor due to its empirical success in various computer vision tasks \cite{XU2025109890}.

\textbf{Consistency Modules.}
To capture both spatial and temporal consistency, we employ transformer blocks inspired by Vision Transformers (ViT) \cite{dosovitskiy2021an}. While convolutional backbones are effective for local feature extraction, they are limited by their fixed receptive fields and inductive biases toward locality. In contrast, transformers naturally model \emph{global dependencies} via self-attention \cite{wang2024comparative}, allowing features from different cameras or time-steps to directly interact regardless of spatial distance. This property is particularly advantageous for adversarial detection: perturbations often appear as localized high-frequency noise, whereas attention mechanisms emphasize coherent global structure.

We therefore introduce two transformer blocks:
\begin{itemize}
    \item The \emph{spatial transformer} models consistency among $[\mathbf{v}_t^{C_\text{left}}, \mathbf{v}_t^{C_\text{centre}}, \mathbf{v}_t^{C_\text{right}}]$, corresponding to the images from the respective cameras, ensuring that overlapping views produce aligned representations. As per the AD systems in our study, we used three cameras for our architecture. However, the spatial transformer module can maintain consistency for more camera inputs.
    \item The \emph{temporal transformer} models consistency across consecutive frames $\mathbf{v}_{t-1}$ and $\mathbf{v}_{t}$, capturing abrupt shifts caused by adversarial perturbations. The transformer block can help capture consistency dependency across a longer time-frame. However, the model was empirically found to perform well with two frames, thus avoiding the increased computational and memory complexity of processing more frames.
\end{itemize}
Each transformer block follows the ViT design: a learnable classification token is appended to the input sequence, combined with positional encodings, and processed through a single layer of multi-head self-attention (four heads). The classification token output provides a compact summary encoding of inter-camera or inter-frame relationships.

\textbf{Classification Head.}
Since our objective is to determine whether the current driving scene is benign or under one of the three adversarial attack categories, $l^{\mathbf{x}_t} \in \{\text{benign}, \text{Poltergeist}, \text{SNAL}, \text{ESIA}\}$, the final stage of the model is naturally formulated as a multi-class classification problem. To this end, we introduce a classification head that maps the aggregated representation from the consistency modules into a prediction score.

By detecting perturbed inputs in real-time, the AD system can issue alarms or take remedial actions such as initiating a safe halt, thereby reducing the risk of catastrophic failures under adversarial conditions. In the next section, we will demonstrate the performance of our proposed detection mechanism.

\section{Empirical Details}
\label{sec4}

Based on the discussions in Section 3, we have divided our empirical evaluation into two parts; discussion on evaluation of existing attacks on end-to-end AD systems, followed by evaluation of \textsc{AD}$^2$ on adversarial attack detection. We experiment in a dynamic simulation environment as it enables us to examine system behaviour under diverse scenarios in a closed loop with inexpensive attack implementations. Similar real world evaluation using a real vehicle would be expensive and demanding to cover various scenarios, and could be taken up as a follow up study.

\subsection{Attack Evaluation on End-to-End AD Systems}

Experiments are conducted in CARLA \cite{dosovitskiyCARLAOpenUrban2017} (v0.9.10.1) with leaderboard 1.0 framework. Simulations run in synchronous mode at 20 Hz. This ensures reliable simulation with better precision. Time-steps are at fixed intervals of 0.05 s (20 Hz) and the simulator waits for the AD agent's response before computing for next time-step. Agents are evaluated on RouteScenario0 of the longest6 benchmark \cite{chittaTransFuserImitationTransformerbased2023}, a 1130m route in Town 1 with high dynamic traffic density. As driving agents, we study two state-of-the-art end-to-end models: Transfuser \cite{chittaTransFuserImitationTransformerbased2023} and Interfuser \cite{shaoSafetyenhancedAutonomousDriving2023}, both pre-trained and publicly released by their corresponding authors. We study their robustness against the Poltergeist \cite{jiPoltergeistAcousticAdversarial2021}, SNAL \cite{chenStealNowAttack2024}, and ESIA \cite{liao2025your} attacks. For evaluation closed-loop performance is measured using route completion ($R$), infraction penalty ($P$), and driving score ($DS$) as in leaderboard 1.0. We additionally report lane deviation $L_{\text{dev}}$ to quantify trajectory stability. Specifically, the metrics can be defined as follows. Detailed definitions and low-level metrics are provided in the Supplementary Material.

%
%

\begin{table*}[h!]
    \centering
    \resizebox{0.7\textwidth}{!}{
    \begin{tabular}{m{2.2cm}|m{2.2cm}| m{1.5cm}| m{1.5cm} m{1.5cm} m{1.5cm}|m{2.6cm}}
    \toprule
    \multirow{2}{6em}{\textbf{AD system}} & \multirow{2}{4em}{\textbf{Attack}} & \textbf{Attack Parameters} & \multicolumn{3}{m{4.5cm}|}{\textbf{Leaderboard Criteria}} &  \textbf{Other Metrics} \\
    \cline{3-3}
    \cline{4-6}
    \cline{7-7}
        &  & Attack interval $d$ &  Driving Score $DS (\downarrow)$ & Infraction Penalty $P (\downarrow)$ & Route Completion $R (\downarrow)$  & Mean $\pm$ std of $|L_{\text{dev}}|$ (m) $(\uparrow)$ \\
    \hline

    \multirow{4}{6em}{Transfuser} & None & None & 100.0 & 1.0 & 100.0 & $0.23\pm0.26$  \\
    \cline{2-7}
     & \multirow{3}{4em}{Poltergeist} & 1 &   0.1263 & 0.0019 & 64.8336 &	$1.47\pm1.21$  \\
     & & 4 & 18.0  & 0.18 & 100.0   & $0.13\pm0.23$ \\
     & & 11 &	 36.0 & 0.36  & 100.0    & $0.22\pm0.26$ \\
    \hline
    \multirow{4}{6em}{Interfuser} & None & None & 74.24 & 1.0 & 74.24 & 0.14 $\pm$ 0.15 \\
    \cline{2-7}
     & \multirow{3}{4em}{Poltergeist} & 1 &  1.47 &	0.16 & 9.16 & 5.13 $\pm$ 3.97 \\
     & & 4 & 26.13 &	0.35 &	74.66  & 0.10 $\pm$ 0.11\\
     & & 11 &	70.0 &	0.7 &	100.0   & 0.19 $\pm$ 0.26\\
    \bottomrule
    
    \end{tabular}
    }
    \caption{Driving performance under poltergeist attack. 
    $d$ is the attack interval, attack at timestep $t$ would be at followed by attack at timestep $t+d$ and so on. 
    For reference, ideal case would have $DS = 100$, $P = 1.0$, and $R = 100\%$ with low $L_{\text{dev}}$. 
    Arrows next to the quantities indicate their respective desired direction of change from attacker's perspective.
    Attack `None' implies baseline.}
    \label{table:poltergeist_attack}
     
\end{table*}

%
%

\begin{table*}[h!]
    \centering
    \resizebox{0.7\textwidth}{!}{
    \begin{tabular}{m{2.2cm}|m{1.4cm}|m{1.5cm} m{1.5cm}|m{1.5cm} m{1.5cm} m{1.5cm}|m{2.5cm}}
    \toprule
    \multirow{2}{6em}{\textbf{AD system}} & \multirow{2}{5em}{\textbf{Attack}} & \multicolumn{2}{m{3cm}|}{\textbf{Attack Parameters}} & \multicolumn{3}{m{4.5cm}|}{\textbf{Leaderboard Criteria}} & \textbf{Other Metrics} \\
    \cline{3-4}
    \cline{5-7}
    \cline{8-8}
        &  & $\epsilon$ ($l_{\infty}$) &  Attack interval $d$ & Driving Score $DS (\downarrow)$ & Infraction Penalty $P (\downarrow)$ & Route Completion $R (\downarrow)$ & Mean $\pm$ std of $|L_{\text{dev}}|$ (m) $(\uparrow)$ \\
    \hline
    \multirow{7}{6em}{Transfuser} & None & None & None & 100.0 & 1.0 & 100.0 & $0.23\pm0.26$  \\
    \cline{2-8}
    & \multirow{6}{4em}{SNAL} 
    & 4 & 1	&   23.4 & 	0.234 &	100.0  &  $0.20 \pm 0.29$ \\												
    & & 4 & 4	&  1.1347 & 0.0298	& 38.0392 & $0.38\pm0.29$ \\
    & & 4 & 11 &  60.0  & 0.6 & 100.0 & $0.14\pm0.24$ \\
    \cline{3-8}
    & & 8 & 1 &   5.4432	&  0.0544 &	100.0 & $0.26\pm0.34$ \\
    & & 8 & 4 &  21.6   &  0.216  & 100.0 & $0.15\pm0.23$\\
    & & 8 & 11 &  60.0  &  0.6  & 100.0 & $0.16\pm0.25$ \\
    \hline
    \multirow{7}{6em}{Interfuser} & None & None & None & 74.24 & 1.0 & 74.24 & $0.14\pm0.15$  \\
    \cline{2-8}
    & \multirow{6}{4em}{SNAL} 
    & 4 & 1	&  70.0 &	0.7 &	100.0  &  $0.11\pm0.11$ \\												
    & & 4 & 4	&    70.0  & 0.7  &	100.0     & $0.16\pm0.15$ \\
    & & 4 & 11 &    42.0   & 0.42 & 100.0     & $0.15\pm0.13$ \\
    \cline{3-8}
    & & 8 & 1 &  50.0	& 0.5 & 100.0 & $0.19\pm0.29$\\
    & & 8 & 4 &   35.0  & 0.35 & 100.0 & $0.13\pm0.13$\\
    & & 8 & 11 &    42.0 & 0.42 & 100.0 & $0.10\pm0.12$ \\
    \bottomrule
    \end{tabular}
    }
    \caption{Driving performance under SNAL. 
    $\epsilon$ is the perturbation bound. 
    $d$ is the attack interval, attack at timestep $t$ would be at followed by attack at timestep $t+d$ and so on. 
    For reference, ideal case would have $DS = 100$, $P = 1.0$, and $R = 100\%$ with low $L_{\text{dev}}$. 
    Arrows next to the quantities indicate their respective desired direction of change from attacker's perspective.
    Attack `None' implies baseline.}
    \label{table:snal_attack}
\end{table*}

%
%
\begin{table*}[h!]
    \centering
    \resizebox{0.70\textwidth}{!}{
    \begin{tabular}{m{2.2cm}|m{1.4cm}|m{1.5cm} m{1.5cm}|m{1.5cm} m{1.5cm} m{1.5cm}|m{2.5cm}}
    \toprule
    \multirow{2}{6em}{\textbf{AD system}} & \multirow{2}{5em}{\textbf{Attack}} & \multicolumn{2}{m{3cm}|}{\textbf{Attack Parameters}} & \multicolumn{3}{m{4.5cm}|}{\textbf{Leaderboard Criteria}} & \textbf{Other Metrics} \\
    \cline{3-4}
    \cline{5-7}
    \cline{8-8}
        &  & Severity &  Attack interval $d$ & Driving Score $DS (\downarrow)$ & Infraction Penalty $P (\downarrow)$ & Route Completion $R (\downarrow)$ & Mean $\pm$ std of $|L_{\text{dev}}|$ (m) $(\uparrow)$ \\
    \hline
    \multirow{10}{6em}{Transfuser} & None & None & None & 100.0 & 1.0 & 100.0 & $0.23\pm0.26$  \\
    \cline{2-8}
    & \multirow{9}{4em}{ESIA} 
    & low & 1	& 12.96	 & 0.1296 & 100.0 &  $0.29\pm0.29$ \\												
    & & low & 4	&  12.96 & 0.1296 & 100.0 & $0.16\pm0.19$ \\
    & & low & 11 &  60.0	& 0.6 & 100.0 & $0.21\pm0.27$ \\
    \cline{3-8}
    & & med & 1 &  69.89 & 1.0 &	69.89 & $0.09\pm0.16$ \\
    & & med & 4 &  100.0 & 1.0 & 100.0 & $0.18\pm0.25$ \\
    & & med & 11 &  21.6 & 0.216 & 100.0 & $0.18\pm0.26$ \\
    \cline{3-8}
    & & high & 1 &   0.263 & 1.0 & 0.263 & $1.04\pm0.72$ \\
    & & high & 4 &   100.0 & 1.0 & 100.0 & $0.17\pm0.25$ \\
    & & high & 11 &  21.6 & 0.216 & 100.0 & $0.19\pm0.21$ \\
    \hline
    \multirow{10}{6em}{Interfuser} & None & None & None & 74.24 & 1.0 & 74.24 & $0.14\pm0.15$  \\
    \cline{2-8}
    & \multirow{9}{4em}{ESIA} 
    & low & 1	&   70.0 & 0.7 & 100.0  &  $0.12\pm0.11$ \\												
    & & low & 4	&   100.0 &	1.0 & 100.0  & $0.09\pm0.11$ \\
    & & low & 11 &    60.0 &	0.6 &	100.0 & $0.18\pm0.15$ \\
    \cline{3-8}
    & & med & 1 &   100.0 &	1.0 &	100.0 & $0.19\pm0.18$ \\
    & & med & 4 &   42.0 &	0.42 & 100.0 & $0.14\pm0.11$ \\
    & & med & 11 &  70.0 & 0.7 & 100.0 & $0.09\pm0.11$ \\
    \cline{3-8}
    & & high & 1 &  60.0 &	0.6 &	100.0 & $0.11\pm0.11$ \\
    & & high & 4 &   12.6 &	0.126 &	100.0 & $0.26\pm0.32$ \\
    & & high & 11 &  100.0 & 1.0 &	100.0 & $0.11\pm0.11$ \\
    \bottomrule
    \end{tabular}
    }
    \caption{Driving performance under ESIA.  
   $d$ is the attack interval.  
    Arrows are from attacker's perspective.
    Attack `None' implies baseline.}
    \label{table:esia_attack}
\end{table*}

\begin{figure*}[h!]
     \centering
     \begin{subfigure}[b]{0.21\textwidth}
         \centering
         \includegraphics[width=\textwidth]{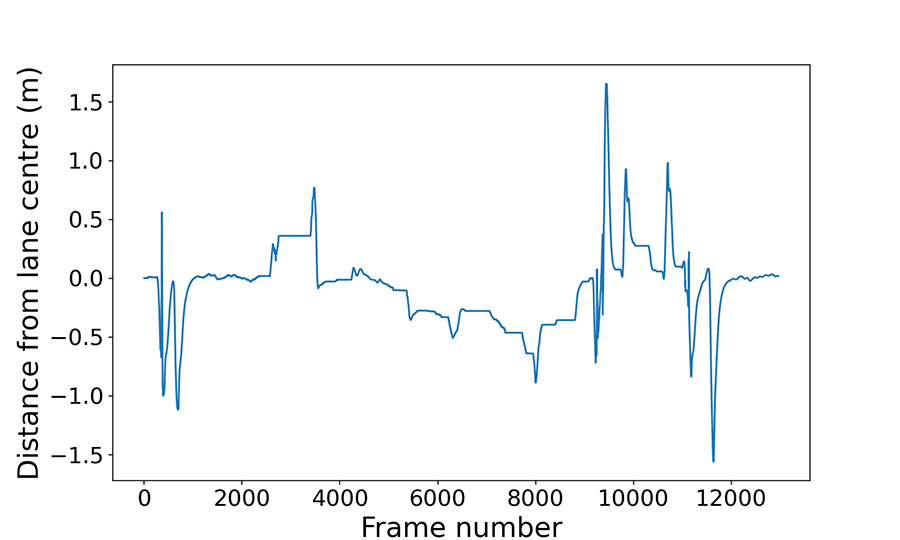}
         \caption{Transfuser: No attack.}
         \label{plot:transfuser_normal}
     \end{subfigure}
     \hfill
     \begin{subfigure}[b]{0.21\textwidth}
         \centering
         \includegraphics[width=\textwidth]{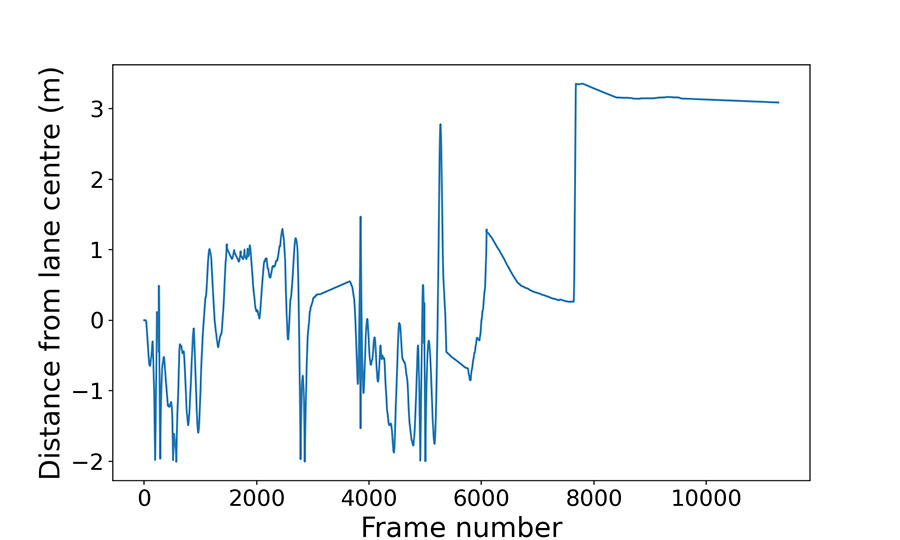}
         \caption{Transfuser: Poltergeist.}
         \label{plot:transfuser_polter}
     \end{subfigure}
     \hfill
     \begin{subfigure}[b]{0.21\textwidth}
         \centering
         \includegraphics[width=\textwidth]{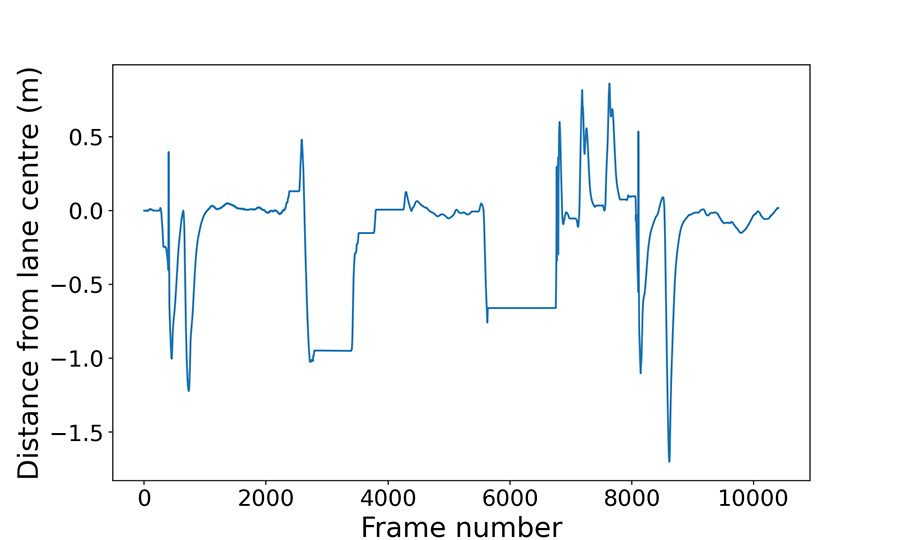}
         \caption{Transfuser: SNAL, $\epsilon=8$.}
         \label{plot:transfuser_snal}
     \end{subfigure}
     \hfill
     \begin{subfigure}[b]{0.21\textwidth}
         \centering
         \includegraphics[width=\textwidth]{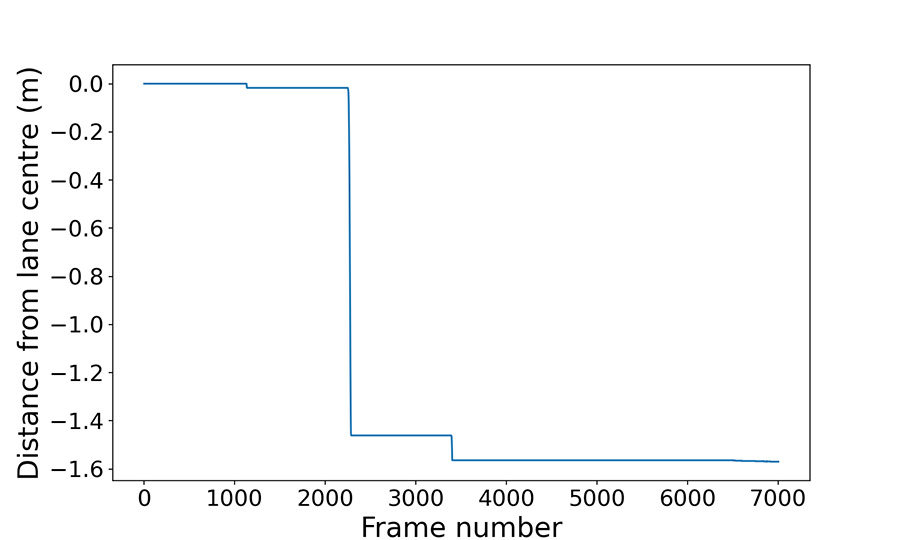}
         \caption{Transfuser: ESIA (high).}
         \label{plot:transfuser_esia}
     \end{subfigure}
     \begin{subfigure}[b]{0.21\textwidth}
         \centering
         \includegraphics[width=\textwidth]{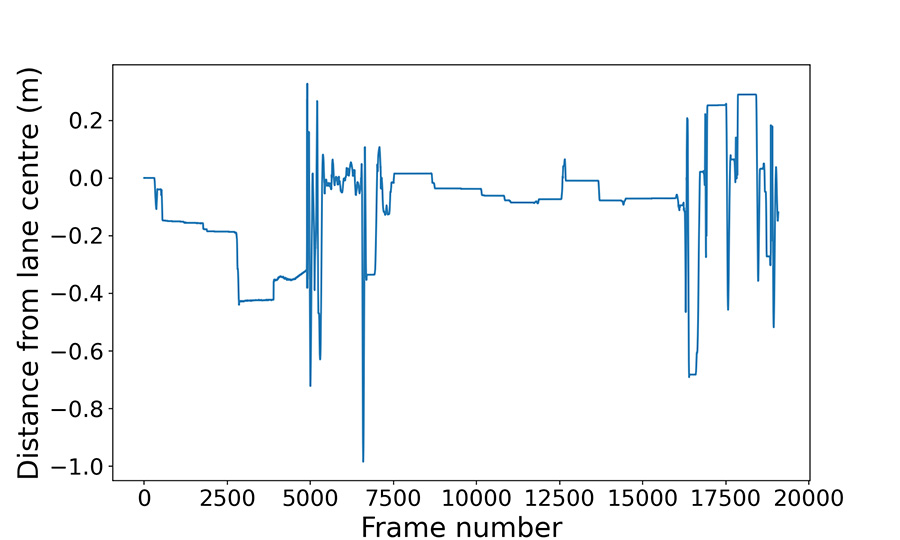}
         \caption{Interfuser: No attack.}
         \label{plot:interfuser_normal}
     \end{subfigure}
     \hfill
     \begin{subfigure}[b]{0.21\textwidth}
         \centering
         \includegraphics[width=\textwidth]{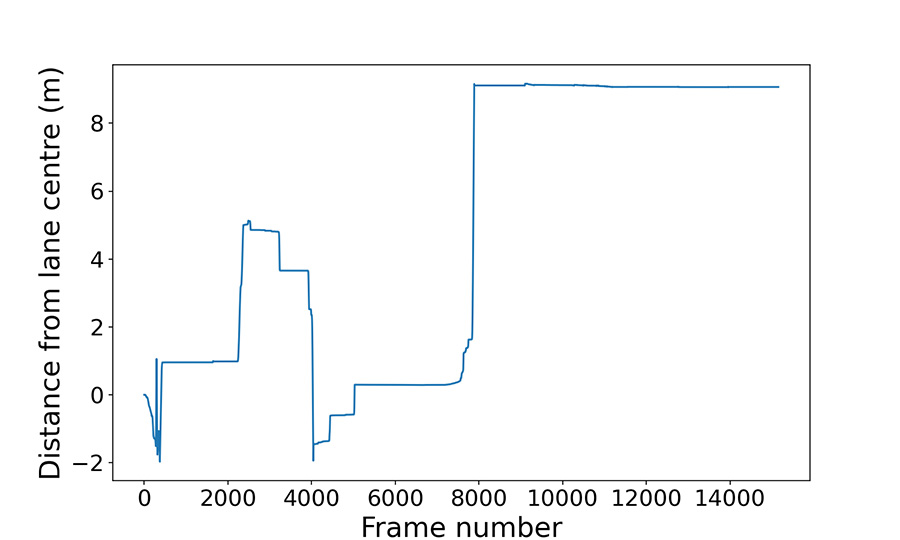}
         \caption{Interfuser: Poltergeist.}
         \label{plot:interfuser_polter}
     \end{subfigure}
     \hfill
     \begin{subfigure}[b]{0.21\textwidth}
         \centering
         \includegraphics[width=\textwidth]{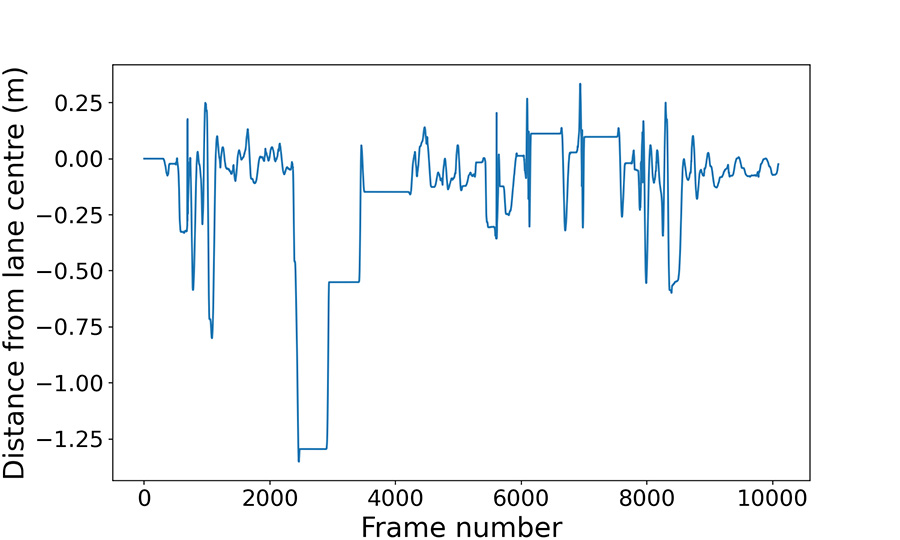}
         \caption{Interfuser: SNAL, $\epsilon=8$.}
         \label{plot:interfuser_snal}
     \end{subfigure}
     \hfill
     \begin{subfigure}[b]{0.21\textwidth}
         \centering
         \includegraphics[width=\textwidth]{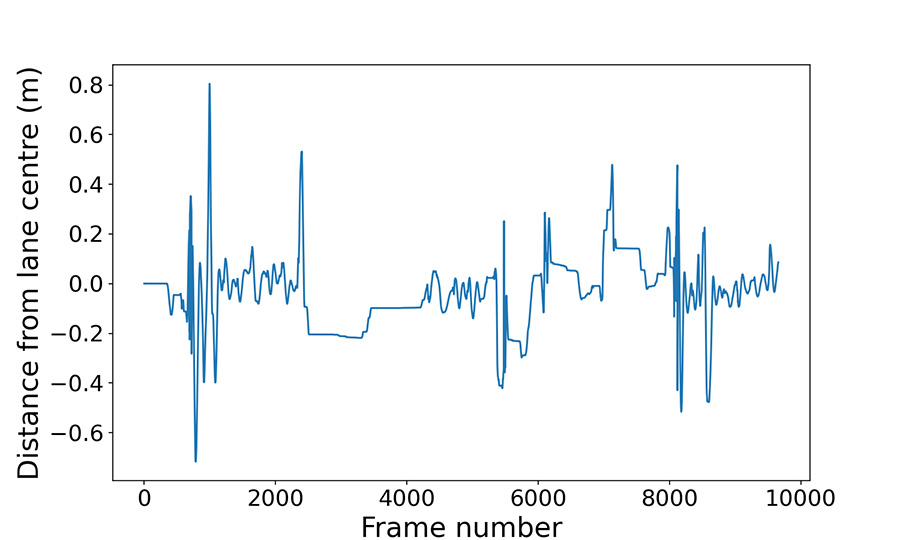}
         \caption{Interfuser: ESIA (high).}
         \label{plot:interfuser_esia}
     \end{subfigure}
        \caption{$L_{\text{dev}}$ plots for AD agents. Attacks are with interval $d=1$.
        Left side of the lane centre is positive. 
        We have different x-axis (Frame number or Timestep) and y-axis (Distance) scales for different configurations. 
        This is due to the different sequence of actions undertaken by the AD system leading to varying driving times. 
        Under strong poltergeist attack on both agents and ESIA on Transfuser, the vehicle deviation is very high. In such cases, the agent goes outside the lane and ultimately gets blocked by some obstacle.
        Deviation under SNAL for both agents and under ESIA for Interfuser are comparatively minor.}
        \label{plot:driving_agents_ldev_plots}
\end{figure*}

\textbf{Route Completion.} 
\textit{
Route completion ($R$) is defined as the percentage of route distance completed by the agent. 
If $l_R$ is the total route distance and $l_A$ is the route distance completed by autonomous driving system $A$ before getting blocked or timed out, then
$R = \frac{l_A}{l_R}\times 100$.
}

\textbf{Infraction Penalty.}
\textit{
Infraction penalty ($P$) is defined as the geometric series of infraction penalty coefficients $p^j$ for every instance of infraction $j$ incurred by the agent during driving.
Different infractions are: collision with pedestrian ($p = 0.5$), collision with other vehicles ($p = 0.6$), collision with static elements ($p = 0.65$), running a red light ($p = 0.7$), running a stop sign ($p = 0.8$).
\begin{equation}
P = \prod_j^{ped,...,stop} (p^j)^{\#infractions}_j
\end{equation}
}

\textbf{Driving Score.}
\textit{
Driving score (DS) is defined as the product of route completion $R$ and infraction penalty $P$.
\begin{equation}
DS = R \times P
\end{equation}
}

\textbf{Deviation from Lane Centre.}
\textit{
Deviation from lane centre ($L_{\text{dev}}$) is the distance between the ego vehicle and the centre of the lane (in metres) at each step or frame of the simulation.
Let $\textbf{l}_{\text{ego}}$ be the vector for location of ego vehicle, $\textbf{l}_{\text{wp}}$ be the corresponding waypoint vector, and $\hat{\textbf{l}}_{\text{wp}\perp}$ be a unit vector perpendicular to $\textbf{l}_{\text{wp}}$.
Then we have, 
\begin{equation}
L_{\text{dev}} = \|(\textbf{l}_{\text{ego}} - \textbf{l}_{\text{wp}})\cdot \hat{\textbf{l}}_{\text{wp}\perp}\|_2
\end{equation} 
}

For Poltergeist, we vary the attack interval $d \in \{1, 4, 11\}$, while for SNAL we additionally control the perturbation bound $\epsilon \in \{4, 8\}$ under the $l_\infty$ norm. We also evaluate ESIA at three levels of severity (low, medium, high). \Cref{table:poltergeist_attack}, \Cref{table:snal_attack}, and \Cref{table:esia_attack} summarize the overall results, while corresponding infraction details are given in the Supplementary Material. Representative lane deviation curves are shown in \Cref{plot:driving_agents_ldev_plots}.

\subsubsection{Findings}
 
Both agents, Transfuser and Interfuser, are highly susceptible to visual perturbations, with driving score reductions of up to 99\% in the worst case. Even intermittent attacks with modest perturbations trigger severe failures such as collisions or red-light violations.\\ 
\noindent
\textbf{Effectiveness of different attacks.}~~
Poltergeist consistently causes stronger degradation compared to $\epsilon$-bounded SNAL and ESIA, largely due to higher-magnitude perturbations that induce visible distortions. They are sufficient to disrupt end-to-end control policies, leading to large lane deviations and unsafe maneuvers. However, such perturbations are easier for human observers to notice. The attack effectiveness of SNAL increases with $\epsilon$-bound. In case of ESIA, trend with respect to severity is not uniform, which may be due to the variance in the positions of colour distortion at critical or non-critical points in the images in different simulation runs.\\ 
\noindent
\textbf{Effect of attack intervals.}~~ 
For Poltergeist, reducing the attack interval monotonically worsens performance, with continuous attacks yielding the most catastrophic outcomes. For SNAL and ESIA, the trend is less regular: performance sometimes degrades even at longer attack intervals, suggesting complex interactions between perturbation timing and environment dynamics. The effectiveness of these attack seems to depend on whether perturbations appear over critical elements in the perceived scene. Importantly, we find that attacks as sparse as twice per second (at 20 Hz control) can destabilize driving leading to hazardous outcomes.\\
\noindent
\textbf{Comparison between agents.}~~ 
There are differences in the architectures of the driving agents; Interfuser also includes a safety controller that applies safety constraints in the prediction of control actions. Therefore the performance degradation was less for Interfuser compared to the Transfuser, owing to the safety controller in Interfuser. However, we also note that this cautious driving with safety constraints resulted in timeout of Interfuser agent under normal conditions (with only 74.24\% of route completed) as it drove slowly with non-essential pre-emptive stops.

These results together highlight that existing end-to-end AD systems remain highly vulnerable to adversarial perception attacks, even under restricted black-box settings. This underlines the need for reliable detection mechanisms.

\begin{table*}[h!]
\centering
\resizebox{\textwidth}{!}{
\begin{tabular}{m{2.2cm}|m{1.4cm}|m{0.9cm} m{1.4cm} m{0.9cm} m{0.9cm} |m{0.9cm} m{1.4cm} m{0.9cm} m{0.9cm} |
    m{0.9cm} m{1.4cm} m{0.9cm} m{0.9cm}}
    \toprule
    \multirow{3}{4em}{\textbf{Method}} & \multirow{3}{4em}{\textbf{Test Accuracy} $(\uparrow)$} & \multicolumn{12}{m{9cm}}{\textbf{Classwise Metrics}} \\
    \cline{3-14}
    & & \multicolumn{4}{m{3.2cm}|}{\textbf{AUC} $(\uparrow)$} & \multicolumn{4}{m{3.6cm}|}{\textbf{TPR} $(\uparrow)$} & 
    \multicolumn{4}{m{3.2cm}}{\textbf{FPR} $(\downarrow)$} \\
    \cline{3-14}
    & & Normal & Poltergeist & SNAL & ESIA & Normal & Poltergeist & SNAL & ESIA & Normal & Poltergeist & SNAL & ESIA\\
    \hline
CoP \cite{fangKernelPCAOutofdistribution2024} & 0.402 &	0.503 & \textbf{1.0} &	0.5 &	0.627 &	0.007 &	\textbf{1.0}	& 0.0 & 1.0 & 0.0 &	\textbf{0.0}	& 0.0 & 0.747 \\
CoRP \cite{fangKernelPCAOutofdistribution2024} & 0.4 & 0.5 & \textbf{1.0} & 0.5 & 0.625 & 0.0 &	\textbf{1.0} &	0.0 &	1.0 &	0.0 &	\textbf{0.0} &	0.0 &	0.75 \\
\hline
LAP4 \cite{pertuzAnalysisFocusMeasure2013} & 0.585 & 0.654 & 0.995 & 0.5 & 0.530 & 0.898 &	\textbf{1.0} &	0.0 &	0.12 &	0.59 &	0.008 &	0.0 &	0.067 \\
\hline
CyberDet \cite{cyberdet} & \underline{0.981} & \underline{0.982}&	\textbf{1.0} &	\underline{0.988} &	\underline{0.979} &	\underline{0.975} &	\textbf{1.0} &	\underline{0.993} &	\underline{0.9573} &	\underline{0.011} &	\textbf{0.0} &	\underline{0.016} &	\textbf{0.0} \\
\hline
\textbf{Our (AD$^2$)} & \textbf{1.0} & \textbf{1.0} &	\textbf{1.0} &	\textbf{1.0} &	\textbf{1.0}	& \textbf{1.0} &	\textbf{1.0} &	\textbf{1.0} &	\textbf{1.0} &	\textbf{0.0} &	\textbf{0.0} &	\textbf{0.0} &	\textbf{0.0}\\
    \bottomrule
\end{tabular}
}
\caption{Performance of different methods in detecting adversarially perturbed
inputs.}
\label{table:detector_offline}
\vspace{-1em}
\end{table*}

\subsection{Evaluation of \textsc{AD}$^2$}

To evaluate the proposed detection model, we construct a dataset with continuous multi-camera image streams in CARLA. The driving agent (Transfuser) is deployed on specified routes from the Longest6 benchmark, and images are recorded at 20Hz. For each sample, we use a pair of three-camera frames at time $t-1$ and $t$, with pairs collected at 1s intervals to ensure scene variability. Following combinations of pairs are present in the dataset: (clean, clean), (clean, attack), (attack, clean), and (attack, attack).  Adversarial perturbations from Poltergeist \cite{jiPoltergeistAcousticAdversarial2021}, SNAL \cite{chenStealNowAttack2024}, and ESIA \cite{liao2025your} are added to generate adversarial data. This yields four classes: benign, Poltergeist, SNAL, and ESIA. For each pair, the class label corresponds to the images at timestep $t$. Train and test splits are kept disjoint by using different routes (RouteScenario19 for training and RouteScenario0 for testing). The class distribution is reported in \Cref{table:dataset_distribution}.
\begin{table}[h!]
\centering
\resizebox{0.45\textwidth}{!}{
\begin{tabular}{m{1.5cm}|m{1cm}|m{1cm}|m{1.4cm}|m{1cm}|m{1cm}}
    \toprule
    \multirow{2}{6em}{\textbf{Data Partition}} & \multirow{2}{6em}{\textbf{Total}} & \multicolumn{4}{m{4cm}}{\textbf{Class Type}} \\
    \cline{3-6}
    & & Benign & Poltergeist & SNAL & ESIA \\
    \midrule
    Train & 11210 & 4484 & 2242 & 2242 & 2242 \\
    Test & 4450 & 1780 & 890 & 890 & 890 \\
    
    \bottomrule
\end{tabular}
}
\caption{Distribution of different classes in the generated dataset.}
\label{table:dataset_distribution}
\end{table}

We report overall accuracy, along with class-wise True Positive Rate (TPR), False Positive Rate (FPR), and Area Under the ROC Curve (AUC). These metrics have been widely used in prior work \cite{carliniAdversarialExamplesAre2017} to evaluate adversarial detectors. We compare against three representative detection baselines: (i) focus measure operators for per pixel focus computation \cite{pertuzAnalysisFocusMeasure2013}, (ii) Kernel PCA based anomaly detection \cite{fangKernelPCAOutofdistribution2024}, and (iii) CyberDet \cite{cyberdet}, a ResNet-based adversarial detector. For fairness, multi-camera inputs are handled consistently across methods. Further details of baseline implementations are provided in the Supplementary Material.

\begin{figure}[h!]
	\centering
    \includegraphics[width=\linewidth]{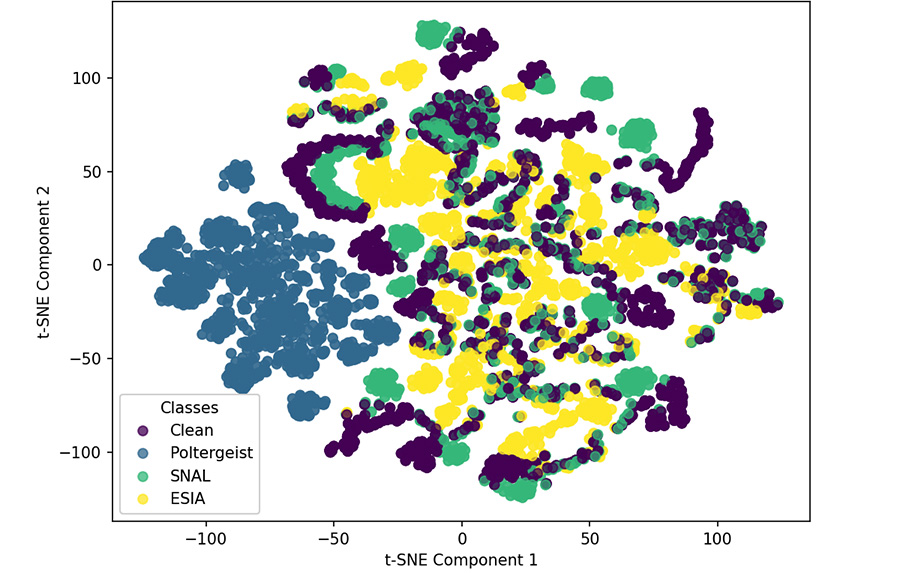}
	\caption{t-SNE visualization of KPCA mapped features. Separation of blur-based poltergeist is visible, but the remaining attacks are enmeshed in distribution of clean benign data.}
	\label{plot:tsne_cop}
    \vspace{-1em}
\end{figure}

\subsubsection{Results}
\noindent
\textbf{OOD Anomaly Analysis.}~~
We first perform an out-of-distribution study using Kernel PCA. As shown in \Cref{plot:tsne_cop}, the blur-based Poltergeist perturbations show partial separation from benign data in the feature space, while SNAL and ESIA remain strongly entangled with clean samples. This suggests that detecting blur-type perturbations is relatively easier, but more subtle adversarial manipulations are harder to isolate without specialized modeling.\\
\noindent
\textbf{Classification Models.}~~
We next evaluate supervised detectors, including the baselines and our proposed model. Quantitative results are summarized in \Cref{table:detector_offline}. We observe that all methods perform well on Poltergeist detection, consistent with the OOD analysis. However, SNAL and ESIA remain challenging, with baselines showing degraded accuracy and more false positives. CyberDet achieves strong detection but at high computational cost. Our proposed model, AD$^2$ achieves consistently superior performance across all classes, with nearly perfect TPR/FPR trade-offs, demonstrating its strong discriminative capability.

We further analyze computational efficiency in \Cref{table:detector_efficiency}. Compared with CyberDet, our model reduces parameter count by nearly 20$\times$ and achieves 1.6$\times$ faster inference on NVIDIA Quadro RTX 5000 GPU. Such improvements are critical for real-time autonomous driving, where detector efficiency directly impacts system responsiveness and safety.

\begin{table}[h!]
\centering
\resizebox{0.40\textwidth}{!}{
\begin{tabular}{m{2.2cm}|m{2.8cm}|m{2.5cm}}
    \toprule
    Method & Model parameters (millions) & Detection time per instance (ms) \\
    \midrule
    CyberDet \cite{cyberdet} & 23.5 M & 0.26 \\
    AD$^2$ &  1.1 M & 0.16 \\
    \bottomrule
\end{tabular}
}
\caption{Comparison of computational efficiency.}
\label{table:detector_efficiency}
\end{table}
\vspace{-1em}
\section{Conclusion}
We presented \textsc{AD}$^2$, a detection model for adversarial signal injection attacks against end-to-end autonomous driving systems. Through closed-loop evaluation in CARLA, a methodology still rarely adopted in adversarial attack research in AD, we analyzed the effect of commonly used attack models on visual perception. Our study shows that even independently trained attackers with perturbations applied to intermittent frames are sufficient to significantly degrade driving performance, showing the vulnerability of state-of-the-art agents. \textsc{AD}$^2$, motivated by recent advances in vision transformers, demonstrates that lightweight transformer blocks coupled with a tailored classification head can provide effective detection capability. While our evaluations are promising, the current study assumes attackers without knowledge of the detection mechanism. Developing feasible adaptive attacks would require very precise acoustic or electromagnetic patterns, which is challenging, given interference from environment noise and background electromagnetic waves. Extending our method to withstand such adaptive strategies, along with assessment of real-world and real-time deployability, and alternate neural architectures are important directions for further research. 
Possible methods for attack mitigation include switching to backup safe mode controllers upon detection of attacks. Such controllers can have safe precomputed actions, like initiating a safe parking etc. This remains part of future work.
Please refer to the Supplementary Material for a discussion on the ethical considerations related to our study.

{
    \small
    \bibliographystyle{ieeenat_fullname}
    \bibliography{main,main_v2}
}

\clearpage
\appendix
\input{appendix}

\end{document}

%% file: appendix.tex
\section{Background}
\subsection*{Perception Systems in Autonomous Driving}
An AD system rely on the information provided by on-board sensors, which allow to describe the state of the vehicle, its environment and other actors \cite{martiReviewSensorTechnologies2019}. 
Perception systems process data from these sensors individually or through data fusion. 
They may include classical sensor data processing algorithms, or machine learning / deep learning models trained for specific perception tasks.
Such systems have the goal to discern both static objects (road and lane markings, road signs, traffic lights, etc.) and dynamic objects (vehicles, pedestrians, etc.) in the environment. 
It is also responsible for its own localization (position, linear and angular velocities, acceleration, orientation).

Cameras are the most common sensors and are available in the market in a wide range of configurations in resolution, frame rate, sensor size, and optics parameters. 
They are low cost and provide a range of information including spatial, dynamic, and semantic. 
They are affected by light and weather conditions.

RADAR (Radio detection and ranging) works using high frequency electromagnetic waves and their reflection from different objects.
Its performance is independent of light and weather conditions \cite{martiReviewSensorTechnologies2019}. 
However, it is affected by reflectivity of different materials. 
Metals amplify radar signals, whereas other materials like wood are virtually transparent.

LIDAR (Light Detection and Ranging) is an active ranging technology that calculates distance to objects by measuring round trip time of a laser pulse \cite{martiReviewSensorTechnologies2019}.
They are useful in creating a highly accurate digital maps using 3D point clouds. 
However, they suffer from several drawbacks: low vertical resolution, sparse measurements with gaps between layers, poor detection of dark and specular objects, and are affected by weather conditions.

IMU (Inertial Measurement Unit) consists of accelerometer to measure linear acceleration, gyroscope to measure orientation and angular velocity, and sometimes a magnetometer for heading reference. 
They are affected by magentic disturbances and time-variant sensor biases and measurement noise \cite{s20216221}.

Speedometer is used to measure the instantaneous speed of the vehicle.

GNSS (Global Naigation Satellite System) refers to any satellite constellation that provides global positioning, navigation, and timing services \cite{WhatGNSSEU}. 
They suffer from atmospheric interference, and availability limitations \cite{joubert2020developments}.

\subsection*{Autonomous Driving Agents in CARLA}
Several autonomous driving agents have been developed and proven on CARLA simulator \cite{dosovitskiyCARLAOpenUrban2017} as part of CARLA Autonomous Driving Leaderboard 1.0 used in CARLA AD challenges.
This leaderboard platform evaluates the driving proficiency of autonomous agents in realistic traffic situations. 

\begin{figure}[h]
	\centering
	\includegraphics[width=.68\linewidth]{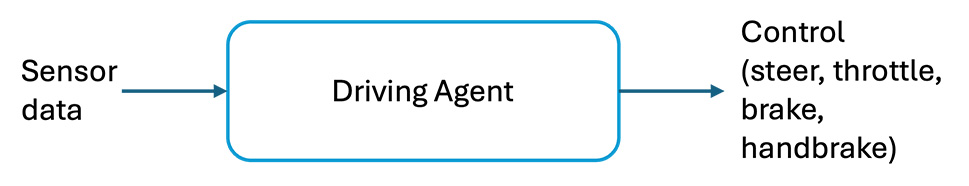}
	\caption{Simplified autonomous driving agent showing expected inputs and outputs.}
	\label{fig:generic}
\end{figure}

All driving agents meant for the leaderboard take sensor data as input and provide control signals such as steer, throttle, brake, handbrake (optional) as output. 
Block diagram representation of a generic AD agent is shown in Figure \ref{fig:generic}.
Internally the driving agent can have different architectures and internal input/output from perception to planning to controller submodules.
Agents can have varying sensor stack within the allowed limits. 
There are individual maximum limits on different sensors -- RGB (Red green blue) camera: 4, LIDAR: 1, RADAR: 2, GNSS: 1, IMU: 1, Speedometer: 1. 
The position of these sensors on the ego vehicle can also vary. 

We briefly discuss two top ranked end-to-end autonomous driving agent that have been proven in the leaderboard 1.0 benchmark. 
They will be the focus of our adversarial attack evaluation study. 

\begin{figure}[h]
	\centering
	\includegraphics[width=.98\linewidth]{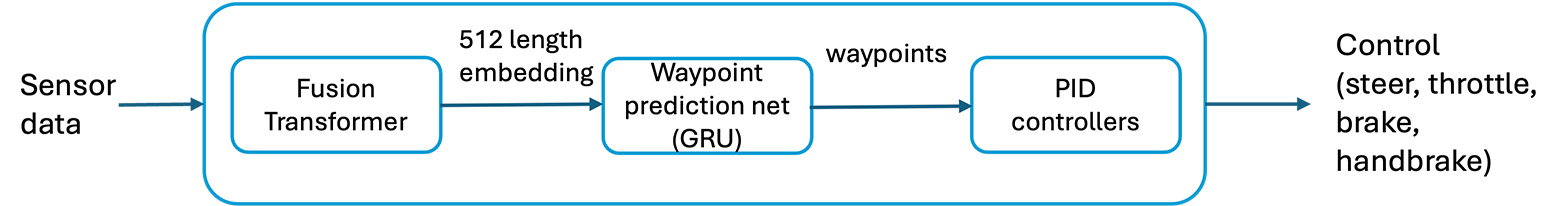}
	\caption{Simplified block diagram for Transfuser agent.}
	\label{fig:transfuser}
\end{figure}

Transfuser \cite{chittaTransFuserImitationTransformerbased2023} uses multi-modal fusion transformer on image and LIDAR inputs to incorporate global context and pairwise interactions into the feature extraction layers.
Using several transformer modules fusion of RGB images and LIDAR representations are performed to yield a 512 dimensional feature vector output. 
This feature vector constitutes a compact representation of the environment that encodes the global context of the scene. 
This is then passed to GRU based waypoint prediction network that predicts the differential waypoints of the ego vehicle. 
These waypoints are then used by the controller to generate steer, throttle and brake values. 
Figure \ref{fig:transfuser} shows a simplified block diagram of their architecture.

\begin{figure}[h]
	\centering
	\includegraphics[width=.98\linewidth]{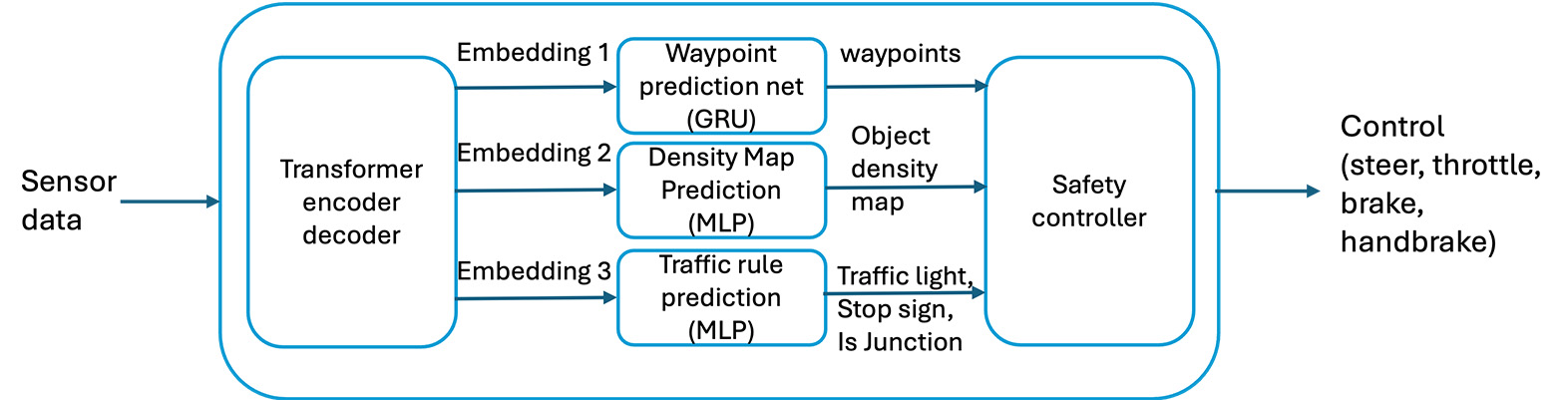}
	\caption{Simplified block diagram for Interfuser agent.}
	\label{fig:interfuser}
\end{figure}

Interfuser \cite{shaoSafetyenhancedAutonomousDriving2023} is an iterpretable sensor fusion transformer, in which information from multi-modal multi-view sensors is fused, which also provides intermediate interpretable features. 
Transformer encoder is used to fuse tokens from different sensors. 
Then, three types of queries are made to the transfomer decoder: waypoint queries, density map queries, and traffic rule query. 
These outputs are then provided to three corresponding prediction headers to predict waypoints, object density map, and traffic rule respectively. 
In the end a safety controller is applied to determine steer, throttle and brake commands. 
Safety controller was designed by the authors to address safety concerns in complex traffic situations. 
This architecture is depicted in Figure \ref{fig:interfuser}.


\subsection*{Evaluation of Autonomous Driving Agents}
Performance of autonomous driving agents and its subsystems can be carried out in real world or through simulation \cite{chenEndtoendAutonomousDriving2024, jia2024benchdrive, Stocco2023}.
In case of real world evaluation, AD agent is deployed in a real vehicle and its capabilities are tested on real physical roads with real obstacles and driving scenarios.
However, this is expensive and may not be feasible for ensuring exhaustive coverage of scenarios. 
Simulation allows validation on almost infinite variability of possible scenarios that could be encountered \cite{Daza2023}. 
Simulation based evaluation can be carried out in different ways as described below:
\begin{itemize}
    \item Open loop (or offline) evaluation: In open loop evaluation, expert trajectories are recorded in datasets with the corresponding sensor data. 
    The agent under evaluation is then used to predict trajectories with the same sensor information which is compared with the expert ground truth \cite{nuScenes}.
    \begin{enumerate}
    \item Model level evaluation: Autonomous driving system may be composed of different subsystems and models. 
    Each model is treated as an independent unit of computation, and it is fed with input data meant for it. 
    The model predicts values that are compared to its specific ground truth labels, which serve as an oracle. 
    For example, testing of object detection model in modular AD systems.
    \item System level evaluation: Overall system is tested on the recorded dataset with expert trajectory as the ground truth. 
    Thus, the individual model's prediction errors become less meaningful.
    As such, failing tests are characterized in terms of the misbehavior of the whole system in response to the AD's predictions. 
    \end{enumerate}
    \item Closed loop (or online) evaluation: Open loop evaluations are insufficient \cite{jia2024benchdrive}. 
    In closed loop evaluation, the driving agent is placed in a dynamic simulation environment. 
    Its predictions and decisions have an immediate effect on the overall system behavior.
\end{itemize}

In our work, we focus on comparison of closed loop performance of the state-of-the-art end-to-end AD systems under adversarial attack with respect to their performance under normal conditions.

\subsection*{Related Work}
We discuss recent work that are closely related to 
our subproblem of adversarial attack evaluation on complete AD
systems. Table \ref{table:lit_comparison} compares our work with such literature.

\begin{table*}[h]
    \centering
    \resizebox{\textwidth}{!}{
    \begin{tabular}{m{1.2cm}|m{2cm}| m{2cm}| m{1.5cm}| m{2cm}| m{2cm}|m{2.4cm}|m{2.4cm}}
    \toprule
    \textbf{Paper} & \textbf{Type of AD System} & \textbf{Sensor configuration of AD System} & \textbf{Sensor under attack} & \textbf{Attack Type} & \textbf{Evaluation Type} & \textbf{Attack strategy} & \textbf{Evaluation Metrics} \\

    \hline
    \cite{dengAnalysisAdversarialAttacks2020} & Simple regression driving models & Camera only & Camera & White-box and black box & Open loop on recorded dataset & Each camera image in the dataset considered individually & Steering angle deviation \\
    \hline
    \cite{wuAdversarialDrivingAttacking2023} & End-to-end & Camera only & Camera & White-box & Closed-loop without traffic & Online attack at each timestep & Steering angle deviation \\ 
    \hline
    \cite{wangAttackEndtoendAutonomous2024} & End-to-end & Camera only & Camera & White-box & Open loop on recorded dataset & Attack optimized for each batch of dataset & L2 error of planned trajectory, collision rate \\

    \hline
    Our work & End-to-end SOTA \cite{chittaTransFuserImitationTransformerbased2023,shaoSafetyenhancedAutonomousDriving2023} & Multimodal & Camera & Restricted black box & Closed loop with traffic & Online continuous attack at each timestep, intermittent timesteps & Driving performance that includes traffic infractions \\
    
    \bottomrule
    
    \end{tabular}
    }
    \caption{Comparison of our adversarial evaluation with those in existing literature. 
    Here we only include work which actually adversarially attack a complete autonomous driving system and analyze the impact.}
    \label{table:lit_comparison}
     
\end{table*}

Few attacks and defense methods on regression-based driving models are analyzed in \cite{dengAnalysisAdversarialAttacks2020}. 
The driving models considered here are simpler. 
They are trained and evaluated offline only on images dataset collected using a front camera. 
Attack goal considered in the paper is also restrictive: attack is considered successful, if the predicted steer angle deviation for a given input frame is greater than specified adversarial threshold, without considering if that actually leads to traffic violations. 

Adversarial attacks on end-to-end AD systems are relatively new and there are limited studies on this topic. 
In \cite{wuAdversarialDrivingAttacking2023}, authors devised white box attacks to deviate the vehicle outside the lane by perturbing the input image. 
They have shown that these attacks are more effective than random noise attacks. 
However, the study was conducted on a simpler end-to-end driving model proposed by the authors themselves which only had camera sensor for perception. 
Simulations were conducted using Udacity and Gazebo simulator. 
The adversary also had complete white box access to the AD system. 
The effectiveness of their attacks on state-of-the-art agents is unknown.
Authors in \cite{wangAttackEndtoendAutonomous2024} propose an attack scheme for end-to-end autonomous driving model through module wise noise injection. 
They assume access to different sub-modules for noise injection as well as for optimization of noise using gradient based methods, that is, complete white box access. 

For completeness and to clarify differentiation, we also briefly mention ANTI-CARLA \cite{ramakrishnaANTICARLAAdversarialTesting2022}. 
This was proposed as an automated testing framework in CARLA for simulating weather conditions (e.g. heavy rain) and sensor faults (e.g. camera occlusion) with the goal of finding driving scenarios that may lead to failure of the system.
This work, however, focuses on generating test cases given some conditions on the enviroment. 
It does not deal with adversarial attacks.

\section{Attack Evaluation on End-to-End AD Systems}
Due to the space constraints in the main paper, we provide additional details on our results here. We first perform the evaluation under normal conditions without any adversarial attack on the AD system and note the baseline performance by monitoring the ego vehicle. 
Subsequently, we evaluate the effect of poltergeist \cite{jiPoltergeistAcousticAdversarial2021}, SNAL \cite{chenStealNowAttack2024}, and ESIA \cite{liao2025your}. 
The time taken by the ego vehicle to complete the same route may differ in different simulation runs. 
Even a slight change in predicted control action of the AD system at any one timestep may cause its own future course and behaviour to change as it also influences the behaviour of surrounding dynamic elements in the environment. 
Also, there is some randomness involved in simulation execution \cite{chanceDeterminismGameEngines2022}. 
We, therefore, focus on macro level indicators of driving performance such as traffic violations and route completion percentage to compute driving score as discussed before. 

In Table 2 of the main paper, we present the degradation in performance of Transfuser and Interfuser AD agents under Poltergeist attack as compared to their performance without any attacks. 
The first row for each agent with `None' adversarial attack is its baseline performance when there is no attack during the entire driving task. 
Then we have rows corresponding to its driving evaluation under poltergeist attack with different attack intervals $d$. 
For each case, we measure driving score $DS$, infraction penalty $P$, route completion $R$, and mean $\pm$ standard deviation of $|L_\text{dev}|$. 
In an ideal case, the AD system would score 100 in $DS$, 1.0 in $P$, and 100 in $R$. 
$L_\text{dev}$, the deviation from lane centre should also be small which would indicate that the vehicle kept to its lane and did not drift out of it. 
In the table, we give mean $\pm$ standard deviation of absolute value $|L_\text{dev}|$ for the entire driving period. 
The arrows next to the quantities indicate their respective desired direction of change from attacker's perspective. 
The attacker would want to prevent the AD vehicle from completing its route (reduce route completion) and increase the number of traffic infractions (decrease in infraction penalty), which would consequently result in lower driving score. Similary, Table 3 and Table 4 in the main paper give the results for corresponding SNAL attack, and ESIA evaluation experiments. 

\begin{table*}[h!]
    \centering
    \resizebox{0.95\textwidth}{!}{
    \begin{tabular}{m{2.1cm}| m{1.2cm}| m{1.3cm}| m{1cm} m{1cm} m{1.1cm} m{1.1cm} m{1.1cm} m{1cm} m{1cm} m{1.2cm} }
    \toprule
    \multirow{2}{6em}{\textbf{AD system}} & \multirow{2}{4em}{\textbf{Attack}} & \textbf{Attack Parameters} & \multicolumn{8}{m{8cm}}{\textbf{Infraction Details}} \\
    \cline{3-3}
    \cline{4-11}
        &  & Attack interval $d$ & Route Completion Test $(\downarrow)$ & Outside Route Lanes Test $(\uparrow)$ & Collision Test $(\uparrow)$ & Running Red Light Test $(\uparrow)$ & Running Stop Sign Test $(\uparrow)$ & In Route Test & Agent Blocked Test & Timeout \\
    \hline

    \multirow{4}{6em}{Transfuser} & None & None & $100\%$ & $0\%$ & $0$ & $0$ & $0$ & S & S & S  \\
    \cline{2-11}
     & \multirow{3}{4em}{Polter-geist} & 1 & $64.83\%$ & $3.16\%$ & $11$ & 3 & 0 &	S &	F &	S  \\
     & & 4 &	100\% & 0\%	& 3 & 0 & 0 & S	& S & S  \\
     & & 11 &	100\% & 0\%	& 2	& 0 & 0 & S & S & S  \\
    \hline

    \multirow{4}{6em}{Interfuser} & None & None & $74.24\%$ & $0\%$	& 0 & 0 & 0	& S & S & F \\
    \cline{2-11}
     & \multirow{3}{4em}{Polter-geist} & 1 & 9.17\%	& 45.41\% & 2 & 1 & 0 &	S &	F &	S  \\
     & & 4 &	74.66\% & 0\% &	1 & 1 & 0 & S &	S &	F \\
     & & 11 &	100\% & 0\%	& 0	& 1 & 0	& S & S & S \\
    \bottomrule
    \end{tabular}
    }
    \caption{Infraction details of the driving agents under poltergeist attack. 
    $d$ is the interval in which attack is carried out. 
    If the first attack was at timestep $t$, then the next attack would be at timestep $t+d$ and so on. 
    S: Success, F: Failure. 
    For reference, ideally route completion should be 100\%, outside route lanes test should be 0\%, different infractions should be 0, and the agent should succeed in the remaining tests.  
    Arrows next to the quantities indicate their respective desired direction of change from attacker's perspective.
    Attack `None' implies baseline.}
    \label{table:poltergeist_criteria}
\end{table*}

\begin{table*}[h!]
    \centering
    \resizebox{0.95\textwidth}{!}{
    \begin{tabular}{m{1.8cm}|m{1.2cm}|m{1cm} m{1cm}|m{1cm} m{1cm} m{1.1cm} m{1.1cm} m{1.1cm} m{1cm} m{1cm} m{1.2cm} }
    \toprule
    \multirow{2}{5em}{\textbf{AD system}} & \multirow{2}{4em}{\textbf{Attack}} & \multicolumn{2}{m{2cm}|}{\textbf{Attack Parameters}} & \multicolumn{8}{m{8cm}}{\textbf{Infraction Details}} \\
    \cline{3-4}
    \cline{5-12}
        &  & $\epsilon$ ($l_{\infty}$) & Attack interval ($d$) & Route Completion Test $(\downarrow)$ & Outside Route Lanes Test $(\uparrow)$ & Collision Test $(\uparrow)$ & Running Red Light Test $(\uparrow)$ & Running Stop Sign Test $(\uparrow)$ & In Route Test & Agent Blocked Test & Timeout \\
    \hline
    \multirow{7}{5em}{Transfuser} & None & None & None & $100\%$ & $0\%$ & $0$ & $0$ & $0$ & S & S & S \\
    \cline{2-12}
    & \multirow{6}{4em}{SNAL} 
    & 4 & 1	& 100\%	& 0\% & 3 & 0 & 0 & S &	S &	S \\												
    & & 4 & 4	& 38.04\% &	1.64\% & 7 & 0 & 0 & S & F & S  \\
    & & 4 & 11 & 100\% & 0\% & 1 & 0 & 0 & S & S & S \\
    \cline{3-12}
    & & 8 & 1 & 100\% & 0\% &	5 &	1 &	0 &	S &	S &	S \\
    & & 8 & 4 & 100\%	& 0\% &	3 & 0 &	0 & S &	S &	S \\
    & & 8 & 11 & 100\%	& 0\% & 1 & 0 & 0 & S &	S & S \\
        \hline
    \multirow{7}{5em}{Interfuser} & None & None & None & $74.24\%$ & $0\%$ & $0$ & $0$ & $0$ & S & S & F \\
    \cline{2-12}
    & \multirow{6}{4em}{SNAL} 
    & 4 & 1	& 100\% & 0\% & 0 & 1 & 0 & S &	S &	S \\												
    & & 4 & 4	&  100\% & 0\% & 0 & 1 & 0  &  S &	S &	S \\
    & & 4 & 11 &  100\%	& 0\% & 1 & 1 & 0  &  S &	S &	S  \\
    \cline{3-12}
    & & 8 & 1 & 100\% & 0\% & 1 & 0 & 0 & S & S & S \\
    & & 8 & 4 &  100\% & 0\% &1 & 1	& 0  &  S &	S &	S  \\
    & & 8 & 11 &  100\% & 0\% &	1 & 1 & 0 &  S & S &	S \\
    \bottomrule
    \end{tabular}
    }
    \caption{Infraction details of the driving agents under SNAL attack. 
    $\epsilon$ is the maximum perturbation that the attacker can introduce.
    $d$ is the interval in which attack is carried out. 
    If the first attack was at timestep $t$, then the next attack would be at timestep $t+d$ and so on. 
    S: Success, F: Failure.  
    For reference, ideally route completion test should be 100\%, outside route lanes test should be 0\%, different infractions should be 0, and the agent should succeed in the remaining tests. 
    Arrows next to the quantities indicate their respective desired direction of change from attacker's perspective.
    Attack `None' implies baseline.}
    \label{table:snal_criteria}
\end{table*}

\begin{table*}[h!]
    \centering
    \resizebox{0.95\textwidth}{!}{
    \begin{tabular}{m{1.8cm}|m{1.2cm}|m{1cm} m{1cm}|m{1cm} m{1cm} m{1.1cm} m{1.1cm} m{1.1cm} m{1cm} m{1cm} m{1.2cm} }
    \toprule
    \multirow{2}{5em}{\textbf{AD system}} & \multirow{2}{4em}{\textbf{Attack}} & \multicolumn{2}{m{2cm}|}{\textbf{Attack Parameters}} & \multicolumn{8}{m{8cm}}{\textbf{Infraction Details}} \\
    \cline{3-4}
    \cline{5-12}
        &  & Severity & Attack interval ($d$) & Route Completion Test $(\downarrow)$ & Outside Route Lanes Test $(\uparrow)$ & Collision Test $(\uparrow)$ & Running Red Light Test $(\uparrow)$ & Running Stop Sign Test $(\uparrow)$ & In Route Test & Agent Blocked Test & Timeout \\
    \hline
    \multirow{10}{5em}{Transfuser} & None & None & None & $100\%$ & $0\%$ & $0$ & $0$ & $0$ & S & S & S \\
    \cline{2-12}
    & \multirow{9}{4em}{ESIA} 
    & low & 1	& 100\%	& 0\% & 4 & 0 & 0 & S &	S &	S \\												
    & & low & 4	& 100\% & 0\% &	4 &	0 &	0 &	S & S & S  \\
    & & low & 11 & 100\% &	0\% & 1	& 0	& 0	 & S & S & S  \\
    \cline{3-12}
    & & med & 1	& 69.9\% & 0\% & 0 & 0 & 0 & S & S & F \\												
    & & med & 4	& 100\%	& 0\% & 0 & 0 & 0 & S & S & S   \\
    & & med & 11 & 100\% & 0\% & 3 & 0 & 0 & S & S & S \\
    \cline{3-12}
    & & high & 1 &	0.26\% & 0\% & 0 & 0 & 0 & S & F & S \\												
    & & high & 4 &	100\% & 0\% & 0 & 0 & 0 & S & S & S  \\
    & & high & 11 & 100\% &	0\%	& 3 & 0	& 0	& S	& S	& S \\
    \hline
    \multirow{10}{5em}{Interfuser} & None & None & None & $74.24\%$ & $0\%$ & $0$ & $0$ & $0$ & S & S & F \\
    \cline{2-12}
    & \multirow{9}{4em}{ESIA} 
    & low & 1	& 100\% & 0\% & 0 & 1 & 0 & S & S & S\\												
    & & low & 4	& 100\% & 0\% &	0 & 0 & 0 &	S & S & S  \\
    & & low & 11 & 100\% & 0\% & 1 & 0 & 0 & S & S & S \\
    \cline{3-12}
    & & med & 1	& 100\%	& 0\% & 0 & 0 & 0 & S & S & S \\												
    & & med & 4	& 100\%	& 0\% & 1 & 1 & 0 & S & S & S  \\
    & & med & 11 & 100\% & 0\% & 0 & 1 & 0 & S & S & S  \\
    \cline{3-12}
    & & high & 1 &	100\% & 0\%	& 1	& 0	& 0	& S	& S	& S \\												
    & & high & 4 &	100\% & 0\%	& 3 & 1 & 0 & S & S & S \\
    & & high & 11 & 100\% & 0\% & 0 & 0 & 0 & S & S & S \\
    \bottomrule
    \end{tabular}
    }
    \caption{Infraction details of the driving agents under ESIA a$d$ is the interval in which attack is carried out. 
    If the first attack was at timestep $t$, then the next attack would be at timestep $t+d$ and so on. 
    S: Success, F: Failure.  
    For reference, ideally route completion test should be 100\%, outside route lanes test should be 0\%, different infractions should be 0, and the agent should succeed in the remaining tests. 
    Arrows next to the quantities indicate their respective desired direction of change from attacker's perspective.
    Attack `None' implies baseline.}
    \label{table:esia_criteria}
\end{table*}

\subsection*{Infraction Details of Adversarial Evaluations}
Table \ref{table:poltergeist_criteria} in this supplementary material provides the details of traffic infractions recorded in each setting along with outcome of few tests. 
Route completion test gives the percentage of route that that agent was able to complete. 
The values in outside route lanes test signifies the percentage of the agent's total driving that was outside its supposed lane in the route. 
The numbers in collision test, running red light test, and running stop sign test gives the count of each of the respective traffic violations that have occured. 
Failure in in-route test implies that the vehicle went off-route by more than 30 m. 
If vehicle controlled by AD system gets blocked by any static or dynamic object for more than 180 s, then it fails the agent blocked test.
Timeout failure happens when the system takes too long to complete the route.

Similary, \Cref{table:snal_criteria} and \Cref{table:esia_criteria} give the results for corresponding SNAL attack and ESIA evaluation experiments. 

\section{Adversarial Attack Detection Baselines}
Adversarially perturbing images introduces artifacts that lead to their detection. 
In order to study the complexity of the detection task we also experiment with several baseline methods. 
This also provides a motivation for our approach.

\noindent \textbf{Focus measure operators} \cite{pertuzAnalysisFocusMeasure2013} \textbf{.}
They are used in the computation of the focus level for every pixel of an image and is the main step in traditional shape-from-focus techniques for recovering 3D shapes \cite{pertuzAnalysisFocusMeasure2013}. 
A variety of algorithms have been proposed in the literature. 
They use different working prinicples: gradients, laplacians, wavelet transforms, image statistics, discrete cosine transforms, etc.
Variance of Laplacian (LAP4) is one of such measure which is defined for point $(i, j)$ in image $I$ as
\begin{equation}
\phi_{i, j} = \sum_{(i, j) \in \Omega(x, y)}(\Delta I(i, j) - \bar{\Delta I})^2
\end{equation}
where $\Delta I$ is the image Laplacian obtained by convolving $I$ with the Laplacian mask,  $\bar{\Delta I}$ is the mean value of the image Laplacian within neigbourhood $\Omega(x, y)$.

In the experiments, laplacian based operators have shown best overall performance at normal imaging conditions but are most sensitive to noise \cite{pertuzAnalysisFocusMeasure2013}. 
They are proven to be very useful in detecting blurs in images.  

\noindent \textbf{Kernel PCA (KPCA)} \cite{fangKernelPCAOutofdistribution2024} \textbf{.} 
Kernel Principal Component Analysis employs non-linear kernels to enhance separability between in-distribution data and out-of-distribution data. 
This has been proposed to overcome the failure of PCA on features obtained using deep neural networks (DNNs).
Given, feature representation $\textbf{z}$ from a DNN, two different feature mappings have been proposed. 
\begin{itemize}
\item Cosine Mapping followed by PCA (CoP): $\phi_{\cos}(\textbf{z})$
\item Cosine and RFFs (Random Fourier Features) Mapping followed by PCA (CoRP): $\phi_{RFF}(\phi_{\cos}(\textbf{z}))$
\end{itemize}
where 
\begin{equation}
\phi_{\cos}(\textbf{z}) = \frac{\textbf{z}}{||\textbf{z}||}
\end{equation}
\begin{equation}
\phi_{RFF}(\textbf{z}) = \sqrt{\frac{2}{M}}[\phi_1(\textbf{z}), \ldots \phi_1(\textbf{z})],
\phi_i(\textbf{z}) = \cos(\textbf{z}^T\omega_i + u_i)
\end{equation}
Here, $\omega_i$ are i.i.d sampled from Fourier transform using kernel $k$, and $u_i$ are i.i.d sampled from Uniform distribution $\mathcal{U}(0, 2\pi)$.
Then, PCA is executed on mapped features for computing the non-linear principal components and corresponding reconstruction errors.
Reconstruction errors have been shown to be very useful in detecting out-of-distribution data.

\noindent \textbf{CyberDet} \cite{cyberdet} \textbf{.} 
CyberDet uses kth order differences computed from the input image fed to a ResNet based binary classifier to discriminate between attacked and genuine images. 
The method is claimed to be agnostic of the attack method,
and of the data used for train or inference.
It has been shown effective against few white-box attacks on the perception system of an autonomous robot.

\section{Potential Negative Societal Impacts}
Like any defense or detection based methods, public knowledge of detection methods may usher in both positive and negative impacts. While, many may benefit from it, potential attackers may exploit that itself further to come up with better attack. Our goal is that we have made it slightly more non-trivial to come up with easy attacks to cause failures in the AD systems.

%% file: main_arxiv.bbl
\begin{thebibliography}{56}
\providecommand{\natexlab}[1]{#1}
\providecommand{\url}[1]{\texttt{#1}}
\expandafter\ifx\csname urlstyle\endcsname\relax
  \providecommand{\doi}[1]{doi: #1}\else
  \providecommand{\doi}{doi: \begingroup \urlstyle{rm}\Url}\fi

\bibitem[Badjie et~al.(2024)Badjie, Cecilio, and Casimiro]{badjieAdversarialAttacksCountermeasures2024}
Bakary Badjie, José Cecilio, and Antonio Casimiro.
\newblock Adversarial attacks and countermeasures on image classification-based deep learning models in autonomous driving systems: a systematic review.
\newblock \emph{Acm Computing Surveys}, 57\penalty0 (1), 2024.
\newblock Number of pages: 52 Place: New York, NY, USA Publisher: Association for Computing Machinery tex.articleno: 20 tex.issue\_date: January 2025.

\bibitem[Caesar et~al.(2020)Caesar, Bankiti, Lang, Vora, Liong, Xu, Krishnan, Pan, Baldan, and Beijbom]{nuScenes}
Holger Caesar, Varun Bankiti, Alex~H. Lang, Sourabh Vora, Venice~Erin Liong, Qiang Xu, Anush Krishnan, Yu Pan, Giancarlo Baldan, and Oscar Beijbom.
\newblock nuscenes: A multimodal dataset for autonomous driving.
\newblock In \emph{2020 IEEE/CVF Conference on Computer Vision and Pattern Recognition (CVPR)}, pages 11618--11628, 2020.

\bibitem[Carlini and Wagner(2017)]{carliniAdversarialExamplesAre2017}
Nicholas Carlini and David Wagner.
\newblock Adversarial examples are not easily detected: {Bypassing} ten detection methods.
\newblock In \emph{Proceedings of the 10th {ACM} workshop on artificial intelligence and security}, pages 3--14, New York, NY, USA, 2017. Association for Computing Machinery.
\newblock Number of pages: 12 Place: Dallas, Texas, USA.

\bibitem[Chance et~al.(2022)Chance, Ghobrial, McAreavey, Lemaignan, Pipe, and Eder]{chanceDeterminismGameEngines2022}
Greg Chance, Abanoub Ghobrial, Kevin McAreavey, Séverin Lemaignan, Tony Pipe, and Kerstin Eder.
\newblock On determinism of game engines used for simulation-based autonomous vehicle verification.
\newblock \emph{IEEE Transactions on Intelligent Transportation Systems}, 23\penalty0 (11):\penalty0 20538--20552, 2022.
\newblock Publisher: IEEE.

\bibitem[Chen et~al.(2024{\natexlab{a}})Chen, Chen, Chung, Lee, and {others}]{chenStealNowAttack2024}
Erh-Chung Chen, Pin-Yu Chen, I Chung, Che-Rung Lee, and {others}.
\newblock Steal now and attack later: {Evaluating} robustness of object detection against black-box adversarial attacks.
\newblock \emph{arXiv preprint arXiv:2404.15881}, 2024{\natexlab{a}}.

\bibitem[Chen et~al.(2024{\natexlab{b}})Chen, Wu, Chitta, Jaeger, Geiger, and Li]{chenEndtoendAutonomousDriving2024}
Li Chen, Penghao Wu, Kashyap Chitta, Bernhard Jaeger, Andreas Geiger, and Hongyang Li.
\newblock End-to-end autonomous driving: {Challenges} and frontiers.
\newblock \emph{IEEE Transactions on Pattern Analysis and Machine Intelligence}, pages 1--20, 2024{\natexlab{b}}.

\bibitem[Chi et~al.(2024)Chi, Msahli, Zhang, Qiu, Zhang, Memmi, and Qiu]{chiAdversarialAttacksAutonomous2024}
Lijun Chi, Mounira Msahli, Qingjie Zhang, Han Qiu, Tianwei Zhang, Gerard Memmi, and Meikang Qiu.
\newblock Adversarial attacks on autonomous driving systems in the physical world: a survey.
\newblock \emph{IEEE Transactions on Intelligent Vehicles}, pages 1--22, 2024.

\bibitem[Chitta et~al.(2023)Chitta, Prakash, Jaeger, Yu, Renz, and Geiger]{chittaTransFuserImitationTransformerbased2023}
Kashyap Chitta, Aditya Prakash, Bernhard Jaeger, Zehao Yu, Katrin Renz, and Andreas Geiger.
\newblock {TransFuser}: {Imitation} with transformer-based sensor fusion for autonomous driving.
\newblock \emph{IEEE Transactions on Pattern Analysis and Machine Intelligence}, 45\penalty0 (11):\penalty0 12878--12895, 2023.

\bibitem[Chougule et~al.(2024)Chougule, Chamola, Sam, Yu, and Sikdar]{chouguleComprehensiveReviewLimitations2024}
Amit Chougule, Vinay Chamola, Aishwarya Sam, Fei~Richard Yu, and Biplab Sikdar.
\newblock A comprehensive review on limitations of autonomous driving and its impact on accidents and collisions.
\newblock \emph{IEEE Open Journal of Vehicular Technology}, 5:\penalty0 142--161, 2024.

\bibitem[Daza et~al.(2023)Daza, Izquierdo, Mart{\'i}nez, Benderius, and Llorca]{Daza2023}
Iv{\'a}n~Garc{\'i}a Daza, Rub{\'e}n Izquierdo, Luis~Miguel Mart{\'i}nez, Ola Benderius, and David~Fern{\'a}ndez Llorca.
\newblock Sim-to-real transfer and reality gap modeling in model predictive control for autonomous driving.
\newblock \emph{Applied Intelligence}, 53\penalty0 (10):\penalty0 12719--12735, 2023.

\bibitem[Deng et~al.(2020)Deng, Zheng, Zhang, Chen, Lou, and Kim]{dengAnalysisAdversarialAttacks2020}
Yao Deng, Xi Zheng, Tianyi Zhang, Chen Chen, Guannan Lou, and Miryung Kim.
\newblock An analysis of adversarial attacks and defenses on autonomous driving models.
\newblock In \emph{2020 {IEEE} international conference on pervasive computing and communications ({PerCom})}, pages 1--10, 2020.

\bibitem[Dosovitskiy et~al.(2017)Dosovitskiy, Ros, Codevilla, Lopez, and Koltun]{dosovitskiyCARLAOpenUrban2017}
Alexey Dosovitskiy, German Ros, Felipe Codevilla, Antonio Lopez, and Vladlen Koltun.
\newblock {CARLA}: {An} open urban driving simulator.
\newblock In \emph{Proceedings of the 1st annual conference on robot learning}, pages 1--16, 2017.

\bibitem[Dosovitskiy et~al.(2021)Dosovitskiy, Beyer, Kolesnikov, Weissenborn, Zhai, Unterthiner, Dehghani, Minderer, Heigold, Gelly, Uszkoreit, and Houlsby]{dosovitskiy2021an}
Alexey Dosovitskiy, Lucas Beyer, Alexander Kolesnikov, Dirk Weissenborn, Xiaohua Zhai, Thomas Unterthiner, Mostafa Dehghani, Matthias Minderer, Georg Heigold, Sylvain Gelly, Jakob Uszkoreit, and Neil Houlsby.
\newblock An image is worth 16x16 words: Transformers for image recognition at scale.
\newblock In \emph{International Conference on Learning Representations}, 2021.

\bibitem[EUSPA(2025)]{WhatGNSSEU}
EUSPA.
\newblock What is {GNSS} {\textbar} {EU} {Agency} for the {Space} {Programme}, 2025.

\bibitem[Fang et~al.(2024)Fang, Tao, Lv, He, Huang, and YANG]{fangKernelPCAOutofdistribution2024}
Kun Fang, Qinghua Tao, Kexin Lv, Mingzhen He, Xiaolin Huang, and JIE YANG.
\newblock Kernel {PCA} for out-of-distribution detection.
\newblock In \emph{The thirty-eighth annual conference on neural information processing systems}, 2024.

\bibitem[Feinman et~al.(2017)Feinman, Curtin, Shintre, and Gardner]{feinmanDetectingAdversarialSamples2017}
Reuben Feinman, Ryan~R. Curtin, Saurabh Shintre, and Andrew~B. Gardner.
\newblock Detecting adversarial samples from artifacts, 2017.
\newblock arXiv: 1703.00410 [stat.ML].

\bibitem[Fingscheidt et~al.(2022)Fingscheidt, Gottschalk, and Houben]{fingscheidtDeepNeuralNetworks2022}
Tim Fingscheidt, Hanno Gottschalk, and Sebastian Houben, editors.
\newblock \emph{Deep {Neural} {Networks} and {Data} for {Automated} {Driving}: {Robustness}, {Uncertainty} {Quantification}, and {Insights} {Towards} {Safety}}.
\newblock Springer International Publishing, Cham, 2022.

\bibitem[Goodfellow et~al.(2015)Goodfellow, Shlens, and Szegedy]{goodfellowExplainingHarnessingAdversarial2015}
Ian~J. Goodfellow, Jonathon Shlens, and Christian Szegedy.
\newblock Explaining and {Harnessing} {Adversarial} {Examples}.
\newblock In \emph{3rd {International} {Conference} on {Learning} {Representations}, {ICLR} 2015, {San} {Diego}, {CA}, {USA}, {May} 7-9, 2015, {Conference} {Track} {Proceedings}}, 2015.

\bibitem[Guesmi and Shafique(2024)]{guesmiNavigatingThreatsSurvey2024}
Amira Guesmi and Muhammad Shafique.
\newblock Navigating threats: a survey of physical adversarial attacks on {LiDAR} perception systems in autonomous vehicles.
\newblock \emph{arXiv preprint arXiv:2409.20426}, 2024.

\bibitem[He et~al.(2016)He, Zhang, Ren, and Sun]{He_2016_CVPR}
Kaiming He, Xiangyu Zhang, Shaoqing Ren, and Jian Sun.
\newblock Deep residual learning for image recognition.
\newblock In \emph{Proceedings of the IEEE Conference on Computer Vision and Pattern Recognition (CVPR)}, 2016.

\bibitem[Ji et~al.(2021)Ji, Cheng, Zhang, Wang, Yan, Xu, and Fu]{jiPoltergeistAcousticAdversarial2021}
Xiaoyu Ji, Yushi Cheng, Yuepeng Zhang, Kai Wang, Chen Yan, Wenyuan Xu, and Kevin Fu.
\newblock Poltergeist: {Acoustic} adversarial machine learning against cameras and computer vision.
\newblock In \emph{2021 {IEEE} symposium on security and privacy ({SP})}, pages 160--175, 2021.

\bibitem[Jia et~al.(2024)Jia, Yang, Li, Zhang, and Yan]{jia2024benchdrive}
Xiaosong Jia, Zhenjie Yang, Qifeng Li, Zhiyuan Zhang, and Junchi Yan.
\newblock Bench2drive: Towards multi-ability benchmarking of closed-loop end-to-end autonomous driving.
\newblock In \emph{The Thirty-eight Conference on Neural Information Processing Systems Datasets and Benchmarks Track}, 2024.

\bibitem[Jiang et~al.(2023)Jiang, Ji, Yan, Xie, Lou, and Xu]{GlitchHiker}
Qinhong Jiang, Xiaoyu Ji, Chen Yan, Zhixin Xie, Haina Lou, and Wenyuan Xu.
\newblock {GlitchHiker}: Uncovering vulnerabilities of image signal transmission with {IEMI}.
\newblock In \emph{32nd USENIX Security Symposium (USENIX Security 23)}, pages 7249--7266, Anaheim, CA, 2023. USENIX Association.

\bibitem[Joubert et~al.(2020)Joubert, Reid, and Noble]{joubert2020developments}
Niels Joubert, Tyler~GR Reid, and Fergus Noble.
\newblock Developments in modern gnss and its impact on autonomous vehicle architectures.
\newblock In \emph{2020 IEEE Intelligent Vehicles Symposium (IV)}, pages 2029--2036. IEEE, 2020.

\bibitem[Kim and Kaur(2024)]{kim2024survey}
Junae Kim and Amardeep Kaur.
\newblock A survey on adversarial robustness of lidar-based machine learning perception in autonomous vehicles.
\newblock \emph{arXiv preprint arXiv:2411.13778}, 2024.

\bibitem[Li et~al.(2025)Li, Wang, Lan, Yu, Wu, and Alvarez]{li2025hydra}
Zhenxin Li, Shihao Wang, Shiyi Lan, Zhiding Yu, Zuxuan Wu, and Jose~M Alvarez.
\newblock Hydra-next: Robust closed-loop driving with open-loop training.
\newblock \emph{arXiv preprint arXiv:2503.12030}, 2025.

\bibitem[Liao et~al.(2025)Liao, Yan, Zhang, Zhai, Wang, and Fu]{liao2025your}
Wenhao Liao, Sineng Yan, Youqian Zhang, Xinwei Zhai, Yuanyuan Wang, and Eugene Fu.
\newblock Is your autonomous vehicle safe? understanding the threat of electromagnetic signal injection attacks on traffic scene perception.
\newblock In \emph{Proceedings of the AAAI Conference on Artificial Intelligence}, pages 27464--27472, 2025.

\bibitem[Liu et~al.(2022)Liu, Levine, Lau, Chellappa, and Feizi]{liuSegmentCompleteDefending2022}
Jiang Liu, Alexander Levine, Chun~Pong Lau, Rama Chellappa, and Soheil Feizi.
\newblock Segment and complete: {{Defending}} object detectors against adversarial patch attacks with robust patch detection.
\newblock In \emph{2022 {{IEEE}}/{{CVF}} Conference on Computer Vision and Pattern Recognition ({{CVPR}})}, pages 14953--14962, Los Alamitos, CA, USA, 2022. IEEE Computer Society.

\bibitem[Madry et~al.(2018)Madry, Makelov, Schmidt, Tsipras, and Vladu]{madryDeepLearningModels2018}
Aleksander Madry, Aleksandar Makelov, Ludwig Schmidt, Dimitris Tsipras, and Adrian Vladu.
\newblock Towards deep learning models resistant to adversarial attacks.
\newblock In \emph{6th international conference on learning representations, {ICLR} 2018, vancouver, {BC}, canada, april 30 - may 3, 2018, conference track proceedings}. OpenReview.net, 2018.
\newblock tex.bibsource: dblp computer science bibliography, https://dblp.org tex.biburl: https://dblp.org/rec/conf/iclr/MadryMSTV18.bib tex.timestamp: Thu, 25 Jul 2019 14:25:44 +0200.

\bibitem[Man et~al.(2023)Man, Muller, Li, Celik, and Gerdes]{manThatPersonMoves2023}
Yanmao Man, Raymond Muller, Ming Li, Z.~Berkay Celik, and Ryan Gerdes.
\newblock That person moves like a car: {{Misclassification}} attack detection for autonomous systems using spatiotemporal consistency.
\newblock In \emph{32nd {{USENIX}} Security Symposium ({{USENIX}} Security 23)}, pages 6929--6946, Anaheim, CA, 2023. USENIX Association.

\bibitem[Marti et~al.(2019)Marti, de~Miguel, Garcia, and Perez]{martiReviewSensorTechnologies2019}
Enrique Marti, Miguel~Angel de Miguel, Fernando Garcia, and Joshue Perez.
\newblock A review of sensor technologies for perception in automated driving.
\newblock \emph{IEEE Intelligent Transportation Systems Magazine}, 11\penalty0 (4):\penalty0 94--108, 2019.

\bibitem[Metzen et~al.(2017)Metzen, Genewein, Fischer, and Bischoff]{metzenDetectingAdversarialPerturbations2017}
Jan~Hendrik Metzen, Tim Genewein, Volker Fischer, and Bastian Bischoff.
\newblock On detecting adversarial perturbations.
\newblock In \emph{International conference on learning representations}, 2017.

\bibitem[Nguyen et~al.(2025)Nguyen, Zhang, Lu, Wu, Zheng, Li~Tan, and Zhen]{nguyenSurveyEvaluationAdversarial2025}
Khoi Nguyen~Tiet Nguyen, Wenyu Zhang, Kangkang Lu, Yu-Huan Wu, Xingjian Zheng, Hui Li~Tan, and Liangli Zhen.
\newblock A survey and evaluation of adversarial attacks in object detection.
\newblock \emph{IEEE Transactions on Neural Networks and Learning Systems}, pages 1--17, 2025.

\bibitem[NHTSA(2017)]{national2017automated}
NHTSA.
\newblock Automated driving systems 2.0: A vision for safety, national highway traffic safety administration and others.
\newblock \emph{Washington, DC: US Department of Transportation, DOT HS}, 812:\penalty0 442, 2017.

\bibitem[Ozaibi et~al.(2024)Ozaibi, Dulva~Hina, and Ramdane-Cherif]{ozaibiEndtoendAutonomousDriving2024}
Youssef~Al Ozaibi, Manolo Dulva~Hina, and Amar Ramdane-Cherif.
\newblock End-to-end autonomous driving in {CARLA}: a survey.
\newblock \emph{IEEE access : practical innovations, open solutions}, 12:\penalty0 146866--146900, 2024.

\bibitem[Pertuz et~al.(2013)Pertuz, Puig, and Garcia]{pertuzAnalysisFocusMeasure2013}
Said Pertuz, Domenec Puig, and Miguel~Angel Garcia.
\newblock Analysis of focus measure operators for shape-from-focus.
\newblock \emph{Pattern Recognition}, 46\penalty0 (5):\penalty0 1415--1432, 2013.

\bibitem[Ramakrishna et~al.(2022)Ramakrishna, Luo, Kuhn, Karsai, and Dubey]{ramakrishnaANTICARLAAdversarialTesting2022}
Shreyas Ramakrishna, Baiting Luo, Christopher~B. Kuhn, Gabor Karsai, and Abhishek Dubey.
\newblock {ANTI}-{CARLA}: {An} adversarial testing framework for autonomous vehicles in {CARLA}.
\newblock In \emph{2022 {IEEE} 25th international conference on intelligent transportation systems ({ITSC})}, pages 2620--2627. IEEE Press, 2022.
\newblock Place: Macau, China Number of pages: 8.

\bibitem[Sasu and Grigorescu(2025)]{cyberdet}
Lucian~M. Sasu and Sorin~M. Grigorescu.
\newblock Cyberdet: Real-time adversarial attacks detection for autonomous robots and self-driving cars.
\newblock In \emph{2025 IEEE Intelligent Vehicles Symposium (IV)}, pages 1935--1941, 2025.

\bibitem[Sato and Chen(2022)]{satoWIPRobustnessLane2022}
Takami Sato and Qi~Alfred Chen.
\newblock {WIP}: {On} robustness of lane detection models to physical-world adversarial attacks in autonomous driving.
\newblock \emph{NDSS 2022 Automotive and Autonomous Vehicle Security (AutoSec) Workshop}, abs/2107.02488, 2022.

\bibitem[Seel et~al.(2020)Seel, Kok, and McGinnis]{s20216221}
Thomas Seel, Manon Kok, and Ryan~S. McGinnis.
\newblock Inertial sensors—applications and challenges in a nutshell.
\newblock \emph{Sensors}, 20\penalty0 (21), 2020.

\bibitem[Shao et~al.(2023{\natexlab{a}})Shao, Wang, Chen, Li, and Liu]{shaoSafetyenhancedAutonomousDriving2023}
Hao Shao, Letian Wang, Ruobing Chen, Hongsheng Li, and Yu Liu.
\newblock Safety-enhanced autonomous driving using interpretable sensor fusion transformer.
\newblock In \emph{Proceedings of the 6th conference on robot learning}, pages 726--737. PMLR, 2023{\natexlab{a}}.

\bibitem[Shao et~al.(2023{\natexlab{b}})Shao, Wang, Chen, Waslander, Li, and Liu]{shaoReasonNetEndtoendDriving2023}
Hao Shao, Letian Wang, Ruobing Chen, Steven~L. Waslander, Hongsheng Li, and Yu Liu.
\newblock {ReasonNet}: {End}-to-end driving with temporal and global reasoning.
\newblock In \emph{Proceedings of the {IEEE}/{CVF} conference on computer vision and pattern recognition ({CVPR})}, pages 13723--13733, 2023{\natexlab{b}}.

\bibitem[Sitawarin et~al.(2018)Sitawarin, Bhagoji, Mosenia, Chiang, and Mittal]{sitawarinDartsDeceivingAutonomous2018}
Chawin Sitawarin, Arjun~Nitin Bhagoji, Arsalan Mosenia, Mung Chiang, and Prateek Mittal.
\newblock Darts: {Deceiving} autonomous cars with toxic signs.
\newblock \emph{arXiv preprint arXiv:1802.06430}, 2018.

\bibitem[Stocco et~al.(2023)Stocco, Pulfer, and Tonella]{Stocco2023}
Andrea Stocco, Brian Pulfer, and Paolo Tonella.
\newblock Model vs system level testing of autonomous driving systems: a replication and extension study.
\newblock \emph{Empirical Software Engineering}, 28\penalty0 (3):\penalty0 73, 2023.

\bibitem[Szegedy et~al.(2014)Szegedy, Zaremba, Sutskever, Bruna, Erhan, Goodfellow, and Fergus]{szegedyIntriguingPropertiesNeural2014}
Christian Szegedy, Wojciech Zaremba, Ilya Sutskever, Joan Bruna, Dumitru Erhan, Ian~J. Goodfellow, and Rob Fergus.
\newblock Intriguing properties of neural networks.
\newblock In \emph{2nd international conference on learning representations, {ICLR} 2014, banff, {AB}, canada, april 14-16, 2014, conference track proceedings}, 2014.
\newblock tex.bibsource: dblp computer science bibliography, https://dblp.org tex.biburl: https://dblp.org/rec/journals/corr/SzegedyZSBEGF13.bib tex.timestamp: Thu, 25 Jul 2019 14:35:25 +0200.

\bibitem[Tramer(2022)]{tramerDetectingAdversarialExamples2022}
Florian Tramer.
\newblock Detecting adversarial examples is ({Nearly}) as hard as classifying them.
\newblock In \emph{Proceedings of the 39th international conference on machine learning}, pages 21692--21702. PMLR, 2022.

\bibitem[Vassilev et~al.(2024)Vassilev, Oprea, Fordyce, and Anderson]{vassilevAdversarialMachineLearning2024}
Apostol Vassilev, Alina Oprea, Alie Fordyce, and Hyrum Anderson.
\newblock Adversarial {Machine} {Learning}: {A} {Taxonomy} and {Terminology} of {Attacks} and {Mitigations}.
\newblock Technical Report NIST Artificial Intelligence (AI) 100-2 E2023, National Institute of Standards and Technology, 2024.

\bibitem[Wang et~al.(2024{\natexlab{a}})Wang, Ma, An, and Dong]{wang2024comparative}
Jing Wang, Siteng Ma, Yu An, and Ruihai Dong.
\newblock A comparative study of vision transformer and convolutional neural network models in geological fault detection.
\newblock \emph{IEEE Access}, 2024{\natexlab{a}}.

\bibitem[Wang et~al.(2024{\natexlab{b}})Wang, Zhang, Han, Fang, Jin, and Kang]{wangAttackEndtoendAutonomous2024}
Lu Wang, Tianyuan Zhang, Yikai Han, Muyang Fang, Ting Jin, and Jiaqi Kang.
\newblock Attack end-to-end autonomous driving through module-wise noise.
\newblock In \emph{Proceedings of the {IEEE}/{CVF} conference on computer vision and pattern recognition ({CVPR}) workshops}, pages 8349--8352, 2024{\natexlab{b}}.

\bibitem[WEF(2025)]{worldeconomicforumAutonomousVehiclesTimeline2025}
WEF.
\newblock Autonomous {Vehicles}: {Timeline} and {Roadmap} {Ahead}, white paper, world economic forum, 2025.

\bibitem[Wu et~al.(2023)Wu, Yunas, Rowlands, Ruan, and Wahlström]{wuAdversarialDrivingAttacking2023}
Han Wu, Syed Yunas, Sareh Rowlands, Wenjie Ruan, and Johan Wahlström.
\newblock Adversarial driving: {Attacking} end-to-end autonomous driving.
\newblock In \emph{2023 {IEEE} intelligent vehicles symposium ({IV})}, pages 1--7, 2023.

\bibitem[Xu et~al.(2025)Xu, Wang, Wu, Leng, and Xu]{XU2025109890}
Guoping Xu, Xiaxia Wang, Xinglong Wu, Xuesong Leng, and Yongchao Xu.
\newblock Development of residual learning in deep neural networks for computer vision: A survey.
\newblock \emph{Engineering Applications of Artificial Intelligence}, 142:\penalty0 109890, 2025.

\bibitem[Xu et~al.(2018)Xu, Evans, and Qi]{xuFeatureSqueezingDetecting2018}
Weilin Xu, David Evans, and Yanjun Qi.
\newblock Feature squeezing: {Detecting} adversarial examples in deep neural networks.
\newblock In \emph{25th annual network and distributed system security symposium, {NDSS} 2018, san diego, california, {USA}, february 18-21, 2018}. The Internet Society, 2018.
\newblock tex.bibsource: dblp computer science bibliography, https://dblp.org tex.timestamp: Thu, 15 Jun 2023 16:53:22 +0200.

\bibitem[Xu et~al.(2024)Xu, Deng, Han, Li, Qiu, and Zhang]{xuPhyScoutDetectingSensor2024}
Yuan Xu, Gelei Deng, Xingshuo Han, Guanlin Li, Han Qiu, and Tianwei Zhang.
\newblock {{PhyScout}}: {{Detecting}} sensor spoofing attacks via spatio-temporal consistency.
\newblock In \emph{Proceedings of the 2024 on {{ACM SIGSAC}} Conference on Computer and Communications Security}, pages 1879--1893, Salt Lake City, UT, USA and New York, NY, USA, 2024. Association for Computing Machinery.

\bibitem[Zhang et~al.(2022)Zhang, Lou, Wang, Wu, Lu, and Jia]{zhangEvaluatingAdversarialAttacks2022}
Jindi Zhang, Yang Lou, Jianping Wang, Kui Wu, Kejie Lu, and Xiaohua Jia.
\newblock Evaluating adversarial attacks on driving safety in vision-based autonomous vehicles.
\newblock \emph{IEEE Internet of Things Journal}, 9\penalty0 (5):\penalty0 3443--3456, 2022.

\bibitem[Zhu et~al.(2023)Zhu, Ji, Cheng, Zhang, and Xu]{tpatch}
Wenjun Zhu, Xiaoyu Ji, Yushi Cheng, Shibo Zhang, and Wenyuan Xu.
\newblock {TPatch}: A triggered physical adversarial patch.
\newblock In \emph{32nd USENIX Security Symposium (USENIX Security 23)}, pages 661--678, Anaheim, CA, 2023. USENIX Association.

\end{thebibliography}
